\documentclass[conference]{IEEEtran}
\IEEEoverridecommandlockouts
\usepackage{cite}
\usepackage{amsmath,amssymb,amsfonts,float}
\usepackage{algorithm,algorithmic}
\usepackage{graphicx}
\usepackage{subfigure}
\usepackage{textcomp}
\usepackage{xcolor}
\usepackage{multirow}
\newtheorem{theorem}{Theorem}
\newtheorem{definition}{Definition}
\def\BibTeX{{\rm B\kern-.05em{\sc i\kern-.025em b}\kern-.08em
    T\kern-.1667em\lower.7ex\hbox{E}\kern-.125emX}}
\begin{document}

\title{Towards Unbiased Random Features with Lower Variance For Stationary Indefinite Kernels\\
\thanks{$^{\star}$Correspondence author: Xiaolin Huang}
}

\author{\IEEEauthorblockN{Qin Luo, Kun Fang, Jie Yang, Xiaolin Huang$^{\star}$}
\IEEEauthorblockA{\textit{Institute of Image Processing and Pattern Recognition} \\
\textit{Shanghai Jiao Tong University}\\
Shanghai, China\\
$\{$tomqin,fanghenshao,jieyang,xiaolinhuang$\}$@sjtu.edu.cn}
}
\maketitle

\begin{abstract}
Random Fourier Features (RFF) demonstrate well-appreciated performance in kernel approximation for large-scale situations but restrict kernels to be stationary and positive definite. And for non-stationary kernels, the corresponding RFF could be converted to that for stationary indefinite kernels when the inputs are restricted to the unit sphere. Numerous methods provide accessible ways to approximate stationary but indefinite kernels. However, they are either biased or possess large variance. In this article, we propose the generalized orthogonal random features, an unbiased estimation with lower variance. Experimental results on various datasets and kernels verify that our algorithm achieves lower variance and approximation error compared with the existing kernel approximation methods. With better approximation to the originally selected kernels, improved classification accuracy and regression ability is obtained with our approximation algorithm in the framework of support vector machine and regression.
\end{abstract}


\section{Introduction}
\label{sec:intro}
Kernel methods are extensively utilized in statistical machine learning with numerous application, such as classification \cite{1995Support, Learningkernels}, regression \cite{1997Support,pmlr-v28-wilson13,Lazaro2010Sparse} and dimensionality reduction \cite{1997Kernel,Carlos2018Convex}. One prominent problem for kernel methods is that it scales poorly with the size of datasets because of higher requirement in computation and storage space. Given $N$ observations, the computational complexity is $\mathcal{O}(N^{3})$ and the storage complexity is $\mathcal{O}(N^{2})$. Random Fourier Features (RFF) \cite{NIPS2007_3182} reduce the computation cost and storage space to $\mathcal{O}(Ns^{2})$ and $\mathcal{O}(Ns)$ through unbiased approximation, where $s$ is the number of Random Fourier Features and much smaller than $N$. However, there exists relatively large variance for initial random features, which is an obstacle for better approximation. Different methods are considered to achieve more accurate and faster approximation for random features, including Fastfood via Hadamard matrices \cite{pmlr-v28-le13}, Quasic Monte Carlo Sampling with a low discrepancy sequence \cite{pmlr-v32-yangb14} and the usage of circulant matrices \cite{Circulant}. Orthogonality is the most effective property to reduce the variance in high-dimensional approximation. Felix et al. \cite{NIPS2016_6246} firstly propose the orthogonality for random features and study the case for Gaussian kernels. Choromanski et al. \cite{pmlr-v84-choromanski18a,pmlr-v97-choromanski19a} extend orthogonality theorem to arbitrary positive definite (PD) radial basis kernel. Structural orthogonal random features \cite{NIPS2016_6246,pmlr-v54-bojarski17a} are further proposed to improve the efficiency for orthogonality.\par
 Random features demonstrate superior performance in kernel approximation, however, two important characteristics are prerequisite for having random features, 1) shift-invariant (also called stationary), $k(\boldsymbol{x},\boldsymbol{y})=k(\boldsymbol{x}-\boldsymbol{y})$, 2) positive definite, ensuring the Fourier transform of kernel as probability measure. However, many kernels might not satisfy the stationary or PD requirements. Typical examples include polynomial kernel\cite{NIPS2000_1790}, neural tangent kernel (NTK) \cite{NIPS2018_8076}, a linear combination of Gaussian kernels (Delta-Gaussian kernel) \cite{pmlr-v80-oglic18a} and the TL1 kernel \cite{Xiaolin2017Classification}. For non-stationary kernels like polynomial kernel, they could be converted to stationary but indefinite kernels when the inputs are restricted to the unit sphere \cite{NIPS2015_5943}. Therefore, the random features for non-stationary kernels or indefinite kernels could be unified with that for stationary indefinite kernels. \par
Because of the wide application of non-stationary or indefinite kernels, a series of works concentrate on
random features for these kernels. For non-stationary kernels, Random Maclaurin \cite{pmlr-v22-kar12} and Tensor Sketch \cite{TensorSketch} are initially proposed for approximating polynomial kernel. Explicit feature maps are further put forward for addictive kernels \cite{2012Efficient} and the Gaussian kernels on the sphere \cite{2017CROification}. However, there exists relatively large variance for these methods to approximate non-stationary kernels. For indefinite kernels, Pennington et al. \cite{NIPS2015_5943} and Liu et al. \cite{Liu8830377} firstly consider the unification for non-stationary kernels and stationary kernels, and approximate stationary indefinite kernels with Gaussian Mixture Model. Nevertheless, Gaussian Mixture Model is a biased approximation. A recent work \cite{Liu2020GeneralizingRF} utilizes generalized measure to achieve unbiased approximation for stationary indefinite kernels but smaller variance is expected to be obtained.\par
In this work, in order to provide an unbiased estimation with lower variance for stationary indefinite kernels, we propose the generalized orthogonal random features (GORF). Orthogonality is introduced to achieve lower variance for unbiased random approximation towards stationary indefinite kernels. We answer the question that how much variance reduction would be obtained through GORF from the theoretical analysis and experimentally demonstrate that GORF achieves variance reduction effect and lower approximation error. Notice that different to imposing orthogonality on two random matrices separately, we achieve orthogonality between two random matrices.\par
The contributions of this work are summarized as follows:\par
$\bullet$ We propose GORF, an unbiased approximation method with lower variance for stationary indefinite kernels, and extend the orthogonality method to the filed of stationary indefinite kernels.\par
$\bullet$ We theoretically examine the unbiasedness of GORF method, and derive the reduced variance after applying orthogonality to random features for indefinite kernels.\par
$\bullet$ We validate our GORF method on two typical indefinite kernels (polynomial kernels on the unit sphere and delta-gaussian kernels) and several datasets, demonstrating that our method achieves lower variance and approximation error. In addtition, under the framework of support vector machine and support vector regression, our method achieves better classification accuracy and regression ability compared with the existing kernel approximation methods.\par

\section{Preliminaries}
\label{sec::prel}

\subsection{Random Fourier Features}
\label{subsec::rff}
This part briefly introduces Bochner's theorem \cite{bochner} and the corresponding Random Fourier Features. Let $\mathcal{D}=\{\boldsymbol{x_{i}}\}_{i=1}^{N}$ be the dataset with $N$ training samples with $\boldsymbol{x}_{i}\in\mathbb{R}^{d}$. Let $k(\cdot,\cdot)$ be the kernel function and $K=[k(\boldsymbol{x}_{i},\boldsymbol{x}_{j})]_{N\times N}$ be the kernel matrix sampled from $\mathcal{D}$. Then the Bochner's theorem is described as follows.
\begin{theorem}
(Bochner's theorem \cite{bochner}) A continuous and stationary function $k:\mathbb{R}^{d}\times\mathbb{R}^{d}\rightarrow{\mathbb{R}}$ is positive definite if and only if it can be represented as
\begin{equation}
\begin{split}
k(\boldsymbol{x}-\boldsymbol{y})&=\int_{\mathbb{R}^{d}}\exp({\rm i}\boldsymbol{w}^T(\boldsymbol{x}-\boldsymbol{y}))p(d\boldsymbol{w})\\
&=\mathbb{E}_{\boldsymbol{w}\sim p(\boldsymbol{w})}[\exp({\rm i}\boldsymbol{w}^T(\boldsymbol{x}-\boldsymbol{y})], \label{Bochner}
\end{split}    
\end{equation}
where $p(\boldsymbol{w})$ is the positive finite measure over $\boldsymbol{w}$, i is the imaginary unit.
\end{theorem}

Bochner's theorem indicates that the Fourier transform of a positive definite function corresponds to a probability measure. Then $s$ random weights could be sampled from the spectral distribution $p(\boldsymbol{w})$, notated as $\{\boldsymbol{w}_{i}\}_{i=1}^{s}$. Monte-Carlo method is utilized to numerically calculate the original Fourier integration.
\begin{equation}
\label{rff}
    k(\boldsymbol{x}-\boldsymbol{y})\approx\frac{1}{s}\sum_{i=1}^{s}\exp(\rm{i}\boldsymbol{w}_{i}^{T}\boldsymbol{x})\exp(\rm{i}\boldsymbol{w}_{i}^{T}\boldsymbol{y})^{*}
\end{equation}

Since $k(.,.)$ is real-valued kernel function, the imaginary part of Equation \eqref{rff} is discarded:
\begin{equation}
    k(\boldsymbol{x}-\boldsymbol{y})\approx\psi(\boldsymbol{x})^T\psi(\boldsymbol{y})
\end{equation}
where $\psi(\boldsymbol{x})=\frac{1}{\sqrt{s}}[\rm{cos}(\boldsymbol{w}_{1}^{T}\boldsymbol{x}),...,\rm{cos}(\boldsymbol{w}_{s}^{T}\boldsymbol{x}),\rm{sin}(\boldsymbol{w}_{1}^{T}\boldsymbol{x}),...,$
$\rm{sin}(\boldsymbol{w}_{s}^{T}\boldsymbol{x})]$. Then $\psi(\boldsymbol{x})$ is the Random Fourier Features for kernel function. Random Fourier Features is unbiased estimation for positive definite kernels, and many works are aimed at improving its approximation quality \cite{pmlr-v28-le13, pmlr-v32-yangb14, NIPS2016_6246, NEURIPS2018_6e923226}.

\subsection{Signed Measure and Jordan Decomposition}
\label{sec::jordan}
Let $\mu:\mathcal{A}\rightarrow{[0,+\infty]}$ be a measure on a set $\Omega$ satisfying $\mu(\phi)=0$ and $\sigma$-additive. If $\mu(\Omega)=1$, it is a probability measure. The Fourier transform for positive definite kernels is probability measure. However, that for stationary indefinite kernel possesses negative components. It is not a nonnegative Borel measure, or even not a measure. To make Monte Carlo sampling applicable to stationary indefinite kernels, a generalized version of measure allowing for negative values needs to be introduced.
\begin{definition}
\label{signedmeasure}
(Signed Measure \cite{Athreya2006MeasureTA}) Let $\Omega$ be some set, $\mathcal{A}$ be a $\sigma$-algebra of subsets on $\Omega$. A signed measure is a function $\mu:\mathcal{A}\rightarrow{[-\infty,+\infty)}$ or $(-\infty,+\infty]$ satisfying $\sigma$-additivity.
\end{definition}

The definition of signed measure extends the traditional measure to a wider range. And Jordan Decomposition below relates the traditionally nonnegative measure with signed measure, which is a solid theoretical foundation for our method.
\begin{theorem}
(Jordan Decomposition \cite{Kubrusly2015EssentialsOM}) Let $\mu$ be a signed measure defined on the $\sigma$-algebra $\mathcal{A}$ as given in Definition \ref{signedmeasure}. There exist two nonnegative measures $\mu_{+}$ and $\mu_{-}$ (one of them is finite measure) such that $\mu=\mu_{+}-\mu_{-}$
\end{theorem}

Note that Jordan Decomposition is not unique. It can be characterized by the Hahn decomposition theorem: the space $\Omega=\Omega_{+}\cup{\Omega_{-}}$, where $\Omega_{+}$ is the positive set where $\mu(S)\geq0$ for all subsets $S \in \Omega_{+}$ and $\Omega_{-}$ is the negative set where $\mu(S)\leq0$ for all subsets $S \in \Omega_{-}$. The total mass is defined as: $||\mu||=||\mu_{+}||+||\mu_{-}||$.

\section{Our Method}
This section specifically illustrates our GORF method for stationary indefinite kernels. Sec. \ref{sec::unbiased} introduces the random features for stationary indefinite kernels and verifies the unbiasedness. Sec. \ref{sec::orthogonality} demonstrates the orthogonality process. Sec. \ref{sec::theory} provides the theoretical derivation for reduced variance after introducing orthogonality.
\subsection{Unbiased Generalized Random Features}
\label{sec::unbiased}
Sec. \ref{sec::jordan} has mentioned the obstacle that Random Fourier Features are applied to stationary indefinite kernel approximation. The spectrum distribution $p(\boldsymbol{w})$ is not a probability measure and we could not directly utilize Monte Carlo sampling. To make use of Monte Carlo sampling, $p(\boldsymbol{w})$ is viewed as signed measure and can be decomposed into the difference of two nonnegative measures via Jordan Decomposition.
\begin{equation}
\begin{split}
\label{decompose}
    &k(\boldsymbol{x}-\boldsymbol{y})=k(\boldsymbol{z})\\
    &=\int_{\mathbb{R}^{d}}\exp({\rm i}\boldsymbol{w}^T\boldsymbol{z})p(d\boldsymbol{w})\\
    &=\int_{\mathbb{R}^{d}}\exp({\rm i}\boldsymbol{w}^T\boldsymbol{z})p_{+}(d\boldsymbol{w})-\int_{\mathbb{R}^{d}}\exp({\rm i}\boldsymbol{\upsilon}^T\boldsymbol{z})p_{-}(d\boldsymbol{\upsilon})\\
    &=||p_{+}||\mathbb{E}_{\boldsymbol{w}\sim \widetilde{p}_{+}}(\exp({\rm i}\boldsymbol{w}^T\boldsymbol{z}))-||p_{-}||\mathbb{E}_{\boldsymbol{\upsilon}\sim \widetilde{p}_{-}}(\exp({\rm i}\boldsymbol{\upsilon}^T\boldsymbol{z})), \\
\end{split}
\end{equation}

\noindent$p_{+}(\boldsymbol{w})$ and $p_{-}(\boldsymbol{\upsilon})$ are two nonnegative finite measures, and $||p_{+}||$ and $||p_{-}||$ are total masses. $\widetilde{p}_{+}$ and $\widetilde{p}_{-}$ are corresponding normalized probability measures, where $\widetilde{p}_{+}=\frac{p_{+}(\boldsymbol{w})}{||p_{+}||}$ and $\widetilde{p}_{-}=\frac{p_{-}(\boldsymbol{\upsilon})}{||p_{-}||}$. Notice that we suppose the stationary kernels mentioned in this paper are radial, where $k(\boldsymbol{z})=k(||\boldsymbol{z}||)$, and accordingly their Fourier transforms are radial, i.e., $p(\boldsymbol{w})=p(||\boldsymbol{w}||)$. Then the calculation of total masses is shown as follows:
\begin{equation}
    ||p_{+}||=\int_{0}^{\infty}|p_{+}(||\boldsymbol{w}||)|dw, \quad ||p_{-}||=\int_{0}^{\infty}|p_{-}(||\boldsymbol{w}||)|dw
\end{equation}

According to Theorem \ref{Bochner}, two expectations in Equation \eqref{decompose} are replaced by two PD function $\widetilde{k}_{+}(\boldsymbol{z})$ and $\widetilde{k}_{-}(\boldsymbol{z})$, and then we obtain $k_{+}(\boldsymbol{z})$ and $k_{-}(\boldsymbol{z})$ after multiplying $||p_{+}||$ and $||p_{-}||$.
\begin{equation}
    \begin{split}
    \label{subsitute}
        k(\boldsymbol{x}-\boldsymbol{y})&=k(\boldsymbol{z})\\
        &=||p_{+}||\widetilde{k}_{+}(\boldsymbol{z})-||p_{-}||\widetilde{k}_{-}(\boldsymbol{z})\\
        &=k_{+}(\boldsymbol{z})-k_{-}(\boldsymbol{z})
    \end{split}
\end{equation}

Define $\phi(\boldsymbol{x})=\frac{1}{\sqrt{s}}[\psi_{1}(\boldsymbol{x}),\psi_{2}(\boldsymbol{x}),...,\psi_{s}(\boldsymbol{x})]^{T}$ with $\psi_{i}(\boldsymbol{x})$ as:
\begin{equation}
\begin{split}
    \psi_{i}(\boldsymbol{x})=[&\sqrt{||p_{+}||}{\rm cos}(\boldsymbol{w}_{i}^{T}\boldsymbol{x}),\sqrt{||p_{+}||}{\rm sin}(\boldsymbol{w}_{i}^{T}\boldsymbol{x}),\\
    &{\rm i}\sqrt{||p_{-}||}{\rm cos}(\boldsymbol{\upsilon}_{i}^{T}\boldsymbol{x}),{\rm i}\sqrt{||p_{-}||}{\rm sin}(\boldsymbol{\upsilon}_{i}^{T}\boldsymbol{x})]. \label{GRFF}
\end{split}
\end{equation}

\noindent where random weights $\{\boldsymbol{w}_{i}\}_{i=1}^{s}$ and $\{\boldsymbol{\upsilon}_{i}\}_{i=1}^{s}$ are independently sampled from $\widetilde{p}_{+}$ and $\widetilde{p}_{-}$ (also called i.i.d sampling). Then the original kernel function could be represented as:
\begin{equation}
    k(\boldsymbol{x}-\boldsymbol{y}) \approx \frac{1}{s}\sum_{i=1}^{s}\langle \psi_{i}(\boldsymbol{x}),\psi_{i}(\boldsymbol{y})\rangle=\phi(\boldsymbol{x})^T\phi(\boldsymbol{y}). \label{apporxi}
\end{equation}

$\phi(\boldsymbol{x})$ is the generalized random features (GRFF) for stationary indefinite kernels. Different from RFF for PD kernels, we construct the random Fourier maps for stationary indefinite kernels in the complex space and RFF for PD kernels is a special case for this random feature map. Generalized random features demonstrate superiority in reducing the computation time and storage space, which drops to $\mathcal{O}(2Ns)$ and $\mathcal{O}(2Ns^{2})$ respectively.

Equation \eqref{decompose} seems to show that any stationary indefinite kernel could be expressed by the difference of two positive definite kernels, but some stationary kernels do not possess this positive decomposition indeed. \cite{NonPositiveKernels} provides a RKKS judgement for the existence of positive decomposition but this simply applies to some intuitive examples: a linear combination of PD kernels. By virtue of measure decomposition of the signed measure, a necessary and sufficient condition to the existence of positive decomposition is derived for more extensive kernels.
\begin{theorem}\cite{Liu2020GeneralizingRF}
Assume that an indefinite kernel is stationary, and its (generalized) Fourier transform is denoted by the measure $\mu$, then $k$ possesses positive decomposition if and only if the total mass of the measure $\mu$ except for the origin 0 is finite, i.e., $||\mu||\leq \infty$.
\end{theorem}

Unbiasedness is the fundamental characteristics in kernel approximation. We examine the unbiasedness of generalized random features as Theorem \ref{bias} demonstrates:
\begin{theorem}
\label{bias}
$K_{\rm GRFF}(\boldsymbol{z})$ is unbiased for all stationary kernels, i.e., $\mathbb{E}(K_{{\rm GRFF}}(\boldsymbol{z}))=k(\boldsymbol{z})$. And the variance is: $Var(K_{GRFF}(\boldsymbol{z})) = \frac{||p_{+}||^{2}}{s}[\frac{1+\widetilde{k}_{+}(2\boldsymbol{z})}{2}-\widetilde{k}_{+}^{2}(\boldsymbol{z})]+\frac{||p_{-}||^{2}}{s}[\frac{1+\widetilde{k}_{-}(2\boldsymbol{z})}{2}-\widetilde{k}_{-}^{2}(\boldsymbol{z})]$ 

\noindent Proof. $\mathbb{E}(K_{{\rm GRFF}}(\boldsymbol{z}))=\mathbb{E}(\sum_{i=1}^{s}\frac{1}{s}||p_{+}||{\rm cos}(\boldsymbol{w}_{i}^{T}\boldsymbol{z})-\sum_{i=1}^{s}\frac{1}{s}||p_{-}||{\rm cos}(\boldsymbol{\upsilon}_{i}^{T}\boldsymbol{z}))=\sum_{i=1}^{s}\frac{||p_{+}||}{s}\mathbb{E}({\rm cos}(\boldsymbol{w}_{i}^{T}\boldsymbol{z}))-\sum_{i=1}^{s}\frac{||p_{-}||}{s}\mathbb{E}({\rm cos}(\boldsymbol{\upsilon}_{i}^{T}\boldsymbol{z}))$.

\noindent Based on the definition of GRFF, $\boldsymbol{w}_{i}\sim\widetilde{p}_{+}(\boldsymbol{w})$, $\boldsymbol{\upsilon}_{i}\sim\widetilde{p}_{-}(\boldsymbol{\upsilon})$ and Bochner's theorem, we get $\mathbb{E}({\rm cos}(\boldsymbol{w}_{i}^{T}\boldsymbol{z}))=\widetilde{k}_{+}(\boldsymbol{z})$, $\mathbb{E}({\rm cos}(\boldsymbol{\upsilon}_{i}^{T}\boldsymbol{z}))=\widetilde{k}_{-}(\boldsymbol{z})$.

\noindent Then $\mathbb{E}(K_{{\rm GRFF}}(\boldsymbol{z}))=\sum_{i=1}^{s}\frac{||p_{+}||}{s}\widetilde{k}_{+}(\boldsymbol{z})-\sum_{i=1}^{s}\frac{||p_{-}||}{s}\widetilde{k}_{-}(\boldsymbol{z})$\\
$=||p_{+}||\widetilde{k}_{+}(\boldsymbol{z})-||p_{-}||\widetilde{k}_{-}(\boldsymbol{z})=k(\boldsymbol{z})$. This verifies that generalized random feature maps is unbiased for all stationary kernels.

\noindent Since $\{\boldsymbol{w}_{i}\}_{i=1}^{s}$ and $\{\boldsymbol{\upsilon}_{i}\}_{i=1}^{s}$ are i.i.d sampled, $Var(K_{\rm GRFF}(\boldsymbol{z}))=Var[\sum_{i=1}^{s}\frac{1}{s}||p_{+}||\rm{cos}(\boldsymbol{w}_{i}^{T}\boldsymbol{z})-\sum_{i=1}^{s}\frac{1}{s}||p_{-}||{\rm cos}(\boldsymbol{\upsilon}_{i}^{T}\boldsymbol{z})]=\frac{||p_{+}||^{2}}{s^{2}}\sum_{i=1}^{s}Var[{\rm cos}(\boldsymbol{w}_{i}^{T}\boldsymbol{z})]+\frac{||p_{-}||^{2}}{s^{2}}\sum_{i=1}^{s}Var[{\rm cos}(\boldsymbol{\upsilon}_{i}^{T}\boldsymbol{z})]=\frac{||p_{+}||^{2}}{s}(\mathbb{E}[{\rm cos}^{2}(\boldsymbol{w}_{1}^{T}\boldsymbol{z})]-\widetilde{k}_{+}^{2}(\boldsymbol{z}))+\frac{||p_{-}||^{2}}{s}(\mathbb{E}[{\rm cos}^{2}(\boldsymbol{\upsilon}_{1}^{T}\boldsymbol{z})]-\widetilde{k}_{-}^{2}(\boldsymbol{z}))=\frac{||p_{+}||^{2}}{s}[\frac{1+\widetilde{k}_{+}(2\boldsymbol{z})}{2}-\widetilde{k}_{+}^{2}(\boldsymbol{z})]+\frac{||p_{-}||^{2}}{s}[\frac{1+\widetilde{k}_{-}(2\boldsymbol{z})}{2}-\widetilde{k}_{-}^{2}(\boldsymbol{z})]$

\end{theorem}

\subsection{Orthogonality}
\label{sec::orthogonality}
Although generalized random feature maps are unbiased, there exists large variance shown in Theorem \ref{bias}. In unbiased approximation, the variance determines the overall approximation error. The sampling method is the main cause for large variance. GORF method utilizes orthogonal sampling, aimed at making random weight vectors not i.i.d sampled from the spectrum and obtaining lower variance.

Let the sampled random weights $\{\boldsymbol{w}_{i}\}_{i=1}^{s}$ and $\{\boldsymbol{\upsilon}_{i}\}_{i=1}^{s}$ constitute the columns of random matrices $W_{\rm pos}=[\boldsymbol{w}_{1},\boldsymbol{w}_{2},...,\boldsymbol{w}_{s}]$ and $W_{\rm neg}=[\boldsymbol{\upsilon}_{1},\boldsymbol{\upsilon}_{2},...,\boldsymbol{\upsilon}_{s}]$. At the beginning of GORF method, we firstly sample $c$ $\ell_{2}$-norm of random weights from the corresponding probability measures $\widetilde{p}_{+}(\boldsymbol{w})$ and $\widetilde{p}_{-}(\boldsymbol{\upsilon})$ separately,
\begin{equation}
||\boldsymbol{w}_{i}||_{2}\sim\widetilde{p}_{+}(||\boldsymbol{w}||),\quad ||\boldsymbol{\upsilon}_{i}||_{2}\sim\widetilde{p}_{-}(||\boldsymbol{\upsilon}||).
\end{equation}

The direction vector of each random weight is sampled from a normalized Gaussian distribution, notated as $\{\boldsymbol{a}_{i}\}_{i=1}^{m}$ and $\{\boldsymbol{b}_{i}\}_{i=1}^{m}$. And they construct the random matrices $M$. Subsequently, QR decomposition is conducted to obtain orthogonal direction vectors (Notice that the size of orthogonal matrix $2m$ takes the maximum value between $2s$ and $2d$ ($2m={\rm max}(2s,2d)$))
\begin{equation}
\begin{split}
&\boldsymbol{a}_{j}\sim \mathcal{N}(\boldsymbol{0},\boldsymbol{I}_{2m}), \quad \boldsymbol{b}_{j}\sim \mathcal{N}(\boldsymbol{0},\boldsymbol{I}_{2m})\\
&M = [\boldsymbol{a}_{1},...,\boldsymbol{a}_{m},\boldsymbol{b}_{1},...,\boldsymbol{b}_{m}]\\
&M^{orth} = {\rm QR}(M).\\
\label{orth}
\end{split}
\end{equation}
\quad We retain the first $d$ rows after obtaining the orthogonal matrix $M^{orth}$, and normalize each columns, notated as $M^{orthn}$. $\ell$2-norm and the corresponding normalized orthogonal direction vector constitute the column vectors for two random matrices $W_{pos}$ and $W_{neg}$,
 \begin{equation}
     \boldsymbol{w}_{i} = ||\boldsymbol{w}_{i}||_{2}M^{orthn}_{i},\quad \boldsymbol{\upsilon}_{i} = ||\boldsymbol{\upsilon}_{i}||_{2}M^{orthn}_{s+i}.
\label{synthesis}
\end{equation}

Algorithm \ref{gorf_construct} specifically illustrates the overall construction of GORF.
\begin{algorithm}
\caption{The Construction of Generalized Orthogonal Random Features}
\label{gorf_construct}
\begin{algorithmic}[1]
\REQUIRE A stationary kernel function $k(\boldsymbol{x},\boldsymbol{y})=k(z)$, $z=||\boldsymbol{x}-\boldsymbol{y}||$, the dimension of training samples $d$, the number of Generalized Orthogonal Random Features $s$
\ENSURE Generalized Orthogonal Random Feature Maps $\phi(\boldsymbol{x})$ such that $    k(\boldsymbol{x}-\boldsymbol{y})\approx\phi(\boldsymbol{x})^T\phi(\boldsymbol{y})$
\STATE Obtain the measure $p(\boldsymbol{w})$ through Fourier transform, and calculate $p_{+}(\boldsymbol{w})$, $p_{-}(\boldsymbol{w})$, $||p_{+}||$, $||p_{-}||$, $\widetilde{p}_{+}(\boldsymbol{w})$, $\widetilde{p}_{-}(\boldsymbol{w})$ through Jordan Decomposition and normalization.
\STATE Individually sample $s$ $\ell$2-norms $||\boldsymbol{w}_{i}||_{2}$ and $||\boldsymbol{\upsilon}_{i}||_{2}$ from $\widetilde{p}_{+}(\boldsymbol{w})$ and $\widetilde{p}_{-}(\boldsymbol{w})$
\STATE Sample $2m$ direction vectors $\boldsymbol{a}_{j}$ and $\boldsymbol{b}_{j}$ from $\mathcal{N}(\boldsymbol{0},\boldsymbol{I}_{2m})$. Use QR decomposition to get orthogonal matrix $M^{orth}$ and $M^{orthn}$.
\STATE Use Equation \eqref{synthesis} to obtain $\boldsymbol{w}_{i}$ and $\boldsymbol{\upsilon}_{i}$ and form the random matrices $W_{\rm pos}$ and $W_{\rm neg}$
\STATE Output the explicit feature mapping $\phi(\boldsymbol{x})$ associated $\psi_{i}(\boldsymbol{x})$ in Equation \eqref{GRFF}
\end{algorithmic}
\end{algorithm}

\subsection{Theoretical Analysis}
\label{sec::theory}
How much variance reduction could be achieved by introducing orthogonal sampling? It is a crucial problem for GORF method. This section concentrates on the theoretical derivation of variation reduction for orthogonal sampling. In order to formulate the variance difference between applying the orthogonal and i.i.d sampling, we firstly review the variance difference between Random Fourier Features approximation and Orthogonal Random Features approximation (ORF) for PD radial kernel.

\begin{theorem}
\cite{pmlr-v84-choromanski18a}
\label{variation-difference-pd}
For a PD radial kernel $k$ on $\mathbb{R}^{d}$ with Fourier measure $p(\boldsymbol{w})$ and $\boldsymbol{x},\boldsymbol{y} \in \mathbb{R}^{d}$, writing $\boldsymbol{z}=\boldsymbol{x}-\boldsymbol{y}$, we have:
\begin{equation}
\begin{split}
G_{k}(\boldsymbol{z}) &= Var(K_{{\rm ORF}}(\boldsymbol{z})) - Var(K_{{\rm RFF}}(\boldsymbol{z}))\\
&= \frac{s-1}{s}\mathbb{E}_{R_{1}}\left[\frac{J_{\frac{d}{2}-1}(R_{1}||\boldsymbol{z}||)\Gamma(d/2)}{(R_{1}||\boldsymbol{z}||/2)^{\frac{d}{2}-1}}\right]^{2}-\\
&\frac{s-1}{s}\mathbb{E}_{R_{1},R_{2}}\left[\frac{J_{\frac{d}{2}-1}(\sqrt{R_{1}^{2}+R_{2}^{2}}||\boldsymbol{z}||)\Gamma(d/2)}{(\sqrt{R_{1}^{2}+R_{2}^{2}}||\boldsymbol{z}||/2)^{\frac{d}{2}-1}}\right],
\end{split}
\end{equation}
where $R_{1},R_{2}\sim p(\boldsymbol{w})$, and $J_{\alpha}$ is the Bessel function of the first kind of degree $\alpha$
\end{theorem}

Based on Theorem \ref{variation-difference-pd}, we formulate the reduced variance after orthogonal sampling for stationary indefinite kernels. 
\begin{theorem}
\label{indefinite_variance}
$K_{\rm GRFF}(\boldsymbol{z})$ and $K_{\rm GORF}(\boldsymbol{z})$ are unbiased approximation with i.i.d sampling and orthogonal sampling for stationary indefinite kernels. For any stationary indefinite kernel $k$ on $\mathbb{R}^{d}$ with Fourier signed measure $p(\boldsymbol{w})$, the corresponding decomposition is $||p_{+}||\widetilde{k}_{+}$ and $||p_{+}||\widetilde{k}_{-}$. Then we have:
\begin{equation}
\label{theo_GORF}
\begin{split}
    Var[K_{{\rm GORF}}(\boldsymbol{z})] - Var[K_{{\rm GRFF}}(\boldsymbol{z})]=&||p_{+}||^{2}G_{\widetilde{k}_{+}}(\boldsymbol{z})+||p_{-}||^{2}\\
    &G_{\widetilde{k}_{-}}(\boldsymbol{z}) +H(\boldsymbol{z}),
\end{split}
\end{equation}

\noindent where $G_{\widetilde{k}_{+}}(\boldsymbol{z})$, $G_{\widetilde{k}_{-}}(\boldsymbol{z})$ are calculated by Theorem \ref{variation-difference-pd}, and $H(\boldsymbol{z})=2||p_{+}||||p_{-}||[\mathbb{E}(a_1)\mathbb{E}(b_1)
    -\mathbb{E}(a_1b_1)]$, $a_{i}={\rm cos}(\boldsymbol{w}_{i}^{T}$ $\boldsymbol{z}), b_{i}={\rm cos}(\boldsymbol{\upsilon}_{i}^{T}\boldsymbol{z})$

\noindent Proof. Since $\{\boldsymbol{w}_{i}\}_{i=1}^{s}$ and $\{\boldsymbol{\upsilon}_{i}\}_{i=1}^{s}$ are not i.i.d sampled and correlate with each other after orthogonalization,
\begin{equation}
\begin{split}
    &Var(K_{\rm GORF}(\boldsymbol{z})) = Var\left[\sum_{i=1}^{s}\frac{1}{s}||p_{+}||a_{i}-\sum_{i=1}^{s}\frac{1}{s}||p_{-}||b_{i}\right]\\
    &=\mathbb{E}\left[\left(\sum_{i=1}^{s}\frac{1}{s}||p_{+}||a_{i}-\sum_{i=1}^{s}\frac{1}{s}||p_{-}||b_{i}\right)^{2}\right]-\mathbb{E}^{2}\left[\sum_{i=1}^{s}\frac{1}{s}||p_{+}||a_{i}-\right.\\
    &\left.\sum_{i=1}^{s}\frac{1}{s}||p_{-}||b_{i}\right]\\
    &=\frac{||p_{+}||^2}{s^2}\sum_{i=1}^{s}\left[\mathbb{E}^{2}(a_i)-\mathbb{E}(a_i^2)\right]+\frac{||p_{-}||^2}{s^2}\sum_{i=1}^{s}\left[\mathbb{E}^{2}(b_i)-\mathbb{E}(b_i^2)\right]\\
    &+\frac{||p_{+}||^2}{s^2}\sum_{i,j=1,i\neq j}^{s}[\mathbb{E}(a_ia_j)-\mathbb{E}(a_i)\mathbb{E}(a_j)]+\frac{||p_{-}||^2}{s^2}\sum_{i,j=1,i\neq j}^{s}\\
    &[\mathbb{E}(b_ib_j)-\mathbb{E}(b_i^2)]+H(\boldsymbol{z})
\end{split}
\end{equation}
Since $\{\boldsymbol{w}_{i}\}_{i=1}^{s}$ and $\{\boldsymbol{\upsilon}_{i}\}_{i=1}^{s}$ are sampled from $\widetilde{p}_{+}(\boldsymbol{w})$ and $\widetilde{p}_{-}(\boldsymbol{w})$, according to Theorem \ref{bias} and \ref{variation-difference-pd}, then
\begin{equation}
\label{simplfy}
\begin{split}
&Var(K_{\rm GORF}(\boldsymbol{z})) = Var[K_{{\rm GRFF}}(\boldsymbol{z})] + ||p_{+}||^{2}G_{\widetilde{k}_{+}}(\boldsymbol{z})+\\
&||p_{-}||^{2}G_{\widetilde{k}_{-}}(\boldsymbol{z}) +H(\boldsymbol{z})
\end{split}
\end{equation}
For $H(\boldsymbol{z})$ in Equation \eqref{simplfy}, we could further simply this term:
\begin{equation}
\begin{split}
    H(\boldsymbol{z})&=\frac{2}{s^2}||p_{+}||||p_{-}||[\mathbb{E}[\sum_{i=1}^{s}a_i]\mathbb{E}[\sum_{j=1}^{s}b_j]-\mathbb{E}[\sum_{i,j=1}^{s}a_ib_j]]\\
    &=2||p_{+}||||p_{-}||[\mathbb{E}(a_1)\mathbb{E}(b_1)
    -\mathbb{E}(a_1b_1)]
\end{split}
\end{equation}
\end{theorem}

Equation \eqref{theo_GORF} divides the variance difference into two parts. The first part $G_{\widetilde{k}_{+}}(\boldsymbol{z})+G_{\widetilde{k}_{-}}(\boldsymbol{z})$ represents the reduced variance when imposing orthogonality on $W_{\rm pos}$ and $W_{\rm neg}$ separately. The second term $H(\boldsymbol{z})$ introduces the reduced variance after making the columns of two random matrices orthogonal with each other. $G_{\widetilde{k}_{+}}(\boldsymbol{z})+G_{\widetilde{k}_{-}}(\boldsymbol{z})+H(\boldsymbol{z})\leq 0$ corresponds to better performance of GORF over GRFF at $\boldsymbol{z}$.

\section{Experiment}
In this section the performance of our GORF method is evaluated, from the aspects of variance reduction, approximation error and performance in specific tasks like classification and regression. All the experiments are conducted on a standard PC with i7-4970 CPU(3.6GHz) and 32G RAM in Matlab\footnote{https://github.com/tomqingo/GORF}.

\subsection{Experimental Setup}
The approximation to the stationary indefinite kernels is the main target in our work. We evaluate our GORF method on two typical stationary indefinite kernels and four datasets in our experiment.
\subsubsection{kernel}
We select two typical stationary indefinite kernels in our research, polynomial kernel on the unit sphere and delta-gaussian kernel.

Polynomial kernel is the mostly used in kernel methods. In order to extend Random Fourier Features to the polynomial kernel for acceleration, the polynomial kernel is converted to a shift-invariant but indefinite kernels when defined on $\mathcal{S}^{d-1}\times \mathcal{S}^{d-1}$ ($||\boldsymbol{x}||_2=||\boldsymbol{y}||_2=1$).
\begin{equation}
    k(\boldsymbol{x}, \boldsymbol{y})=(1-\frac{||\boldsymbol{x}-\boldsymbol{y}||^{2}}{a^2})^{m}=\alpha(q+\langle\boldsymbol{x},\boldsymbol{y}\rangle)^{m}
\end{equation}

\noindent where $q=a^{2}/2-1$, $\alpha=(2/a^2)^m$. The spectrum distribution of polynomial kernel on the sphere is shown as follows:
\begin{equation}
    p(\boldsymbol{w})=\sum_{i=0}^{p}\frac{m!}{(m-i)!}(1-\frac{4}{a^{2}})^{m-i}(\frac{2}{a^2})^{i}(\frac{2}{||\boldsymbol{w}||_{2}})^{\frac{d}{2}+i}J_{\frac{d}{2}+i}(2||\boldsymbol{w}||_{2})
\end{equation}

Delta-gaussian kernel is the linear combination of commonly used Gaussian kernels, with arbitrary coefficients $\{a_{i}\}_{i=1}^{m}$.
\begin{equation}
    k(\boldsymbol{x}-\boldsymbol{y}) = k(\boldsymbol{z}) = \sum_{i=1}^{m}a_i e^{-\frac{||\boldsymbol{z}||^2}{2\sigma_i^2}}
\end{equation}

The corresponding spectrum distribution is shown below:
\begin{equation}
     p(\boldsymbol{w}) = \sum_{i=1}^{m}a_{i}\sigma_{i}^{d}e^{-\frac{\sigma_{i}^2||\boldsymbol{w}||^2}{2}}
\end{equation}

In the implementation, $p_{+}(\boldsymbol{w})=\max(0,p(\boldsymbol{w}))$, $p_{-}(\boldsymbol{w})=\max(0,-p(\boldsymbol{w}))$.

\subsubsection{Datasets}
Four representative benchmark datasets are adpoted. $letter$\footnote{http://archive.ics.uci.edu/ml/datasets.php}, $ijcnn1$\footnote{https://www.csie.ntu.edu.tw/cjlin/libsvmtools/datasets/}, $usps$\footnote{https://www.kaggle.com/bistaumanga/usps-dataset} are the classification datasets with the data quantity and dimension shown in Table \ref{benchmark}. $housing$\footnote{https://archive.ics.uci.edu/ml/machine-learning-databases/housing/} is the ordinary regression dataset for Boston housing prices with 506 samples and 13 input dimensions. 405 samples are randomly selected for training set and the left 101 samples constitute testing set. For all the datasets, all the inputs are normalized to $[0,1]^{d}$.
\begin{table}[htbp]
\caption{BENCHMARK CLASSIFICATION DATASETS}
\label{benchmark}
\begin{center}
\begin{tabular}{p{50pt} c c c c}
\hline
\textbf{Datasets}&\textbf{d}&\textbf{training}&\textbf{testing}\\
\cline{1-4} 
\textit{letter} & 16 & 12000 & 6000\\
\textit{ijcnn1}& 22 & 49990 & 91701\\
\textit{usps}& 256 & 7291& 2007\\
\hline
\end{tabular}
\label{tab1}
\end{center}
\end{table}

\subsection{Variance Reduction}
\label{sec::var_redcut}
Orthogonality is the crucial part in our GORF algorithm aimed at reducing variance. Sec. \ref{sec::theory} has derived the reduced variance after applying orthogonality to generalized random features. Since the expectation in variance reduction formula could not be calculated analytically, we then numerically calculate the reduced variance for polynomial kernels on the unit sphere and delta-gaussian kernel. Monte-Carlo method is utilized to calculate the expectation using 10000 random weights sampled from $\widetilde{p}_{+}(\boldsymbol{w})$ and $\widetilde{p}_{-}(\boldsymbol{\upsilon})$ respectively.
 \begin{figure}[htbp]
    \centering
    \subfigure[polynomial kernel on the unit sphere ($a=3, m=1$)]{
    \centering
    \includegraphics[width=2.5in]{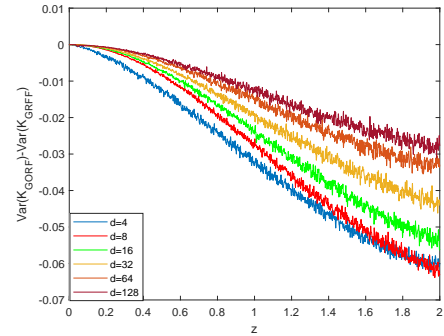}
    }
    \subfigure[delta-gaussian kernel ($a_{1}=1,a_{2}=-1, \sigma_{1}=1, \sigma_{2}=10)$]{
    \centering
    \includegraphics[width=2.5in]{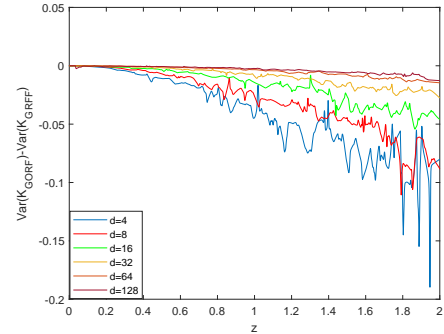}
    }
  \caption{Numerically calculated variance reduction for polynomial kernel on the unit sphere and delta-gaussian kernel under different data dimensions. Observe that the variance difference is negative, orthogonality achieves lower variance for a wide range of mean square distance ($z = ||\boldsymbol{x}-\boldsymbol{y}||_{2}$, $s=d$)}
\end{figure}

Fig. 1 shows the variation reduction for polynomial kernel on the unit sphere and delta-gaussian kernel. The horizontal axis is the mean square distance between two samples. It could be observed that orthogonality has a prominent variance reduction effect for a wide range of mean square distance. Although the variance reduction seems less significant as the data dimension increases in Fig. 1, the relative value for variance reduction remains large for large data dimension, since the variance for i.i.d sampling decreases as the data dimension increases.

\begin{figure*}[htbp]
	\centering
	\subfigure{
		\subfigure{
			\begin{minipage}[t]{0.3\linewidth}
				\centering
				\includegraphics[width=2in]{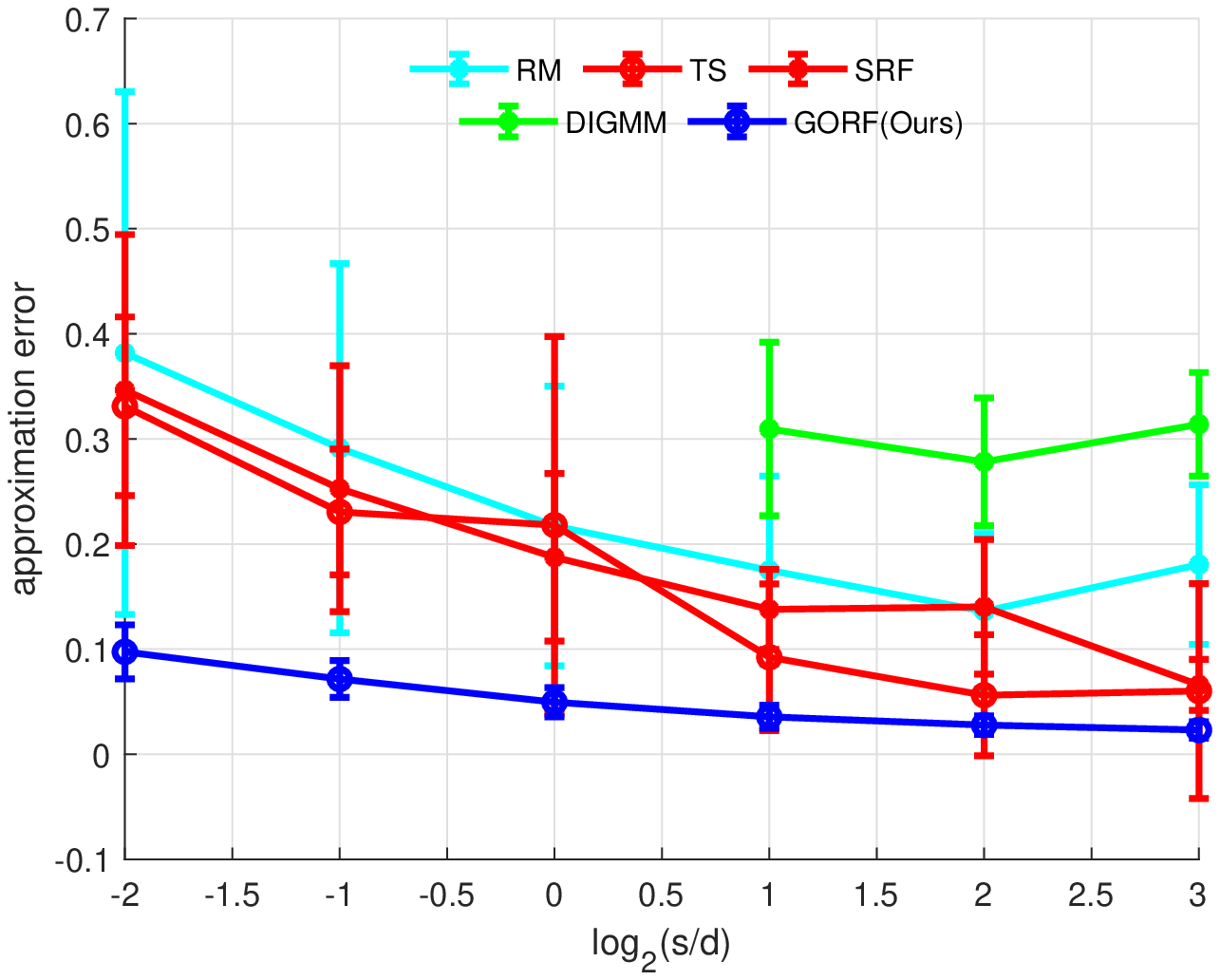}
			\end{minipage}%
		}%
		\subfigure{
			\begin{minipage}[t]{0.3\linewidth}
				\centering
				\includegraphics[width=2in]{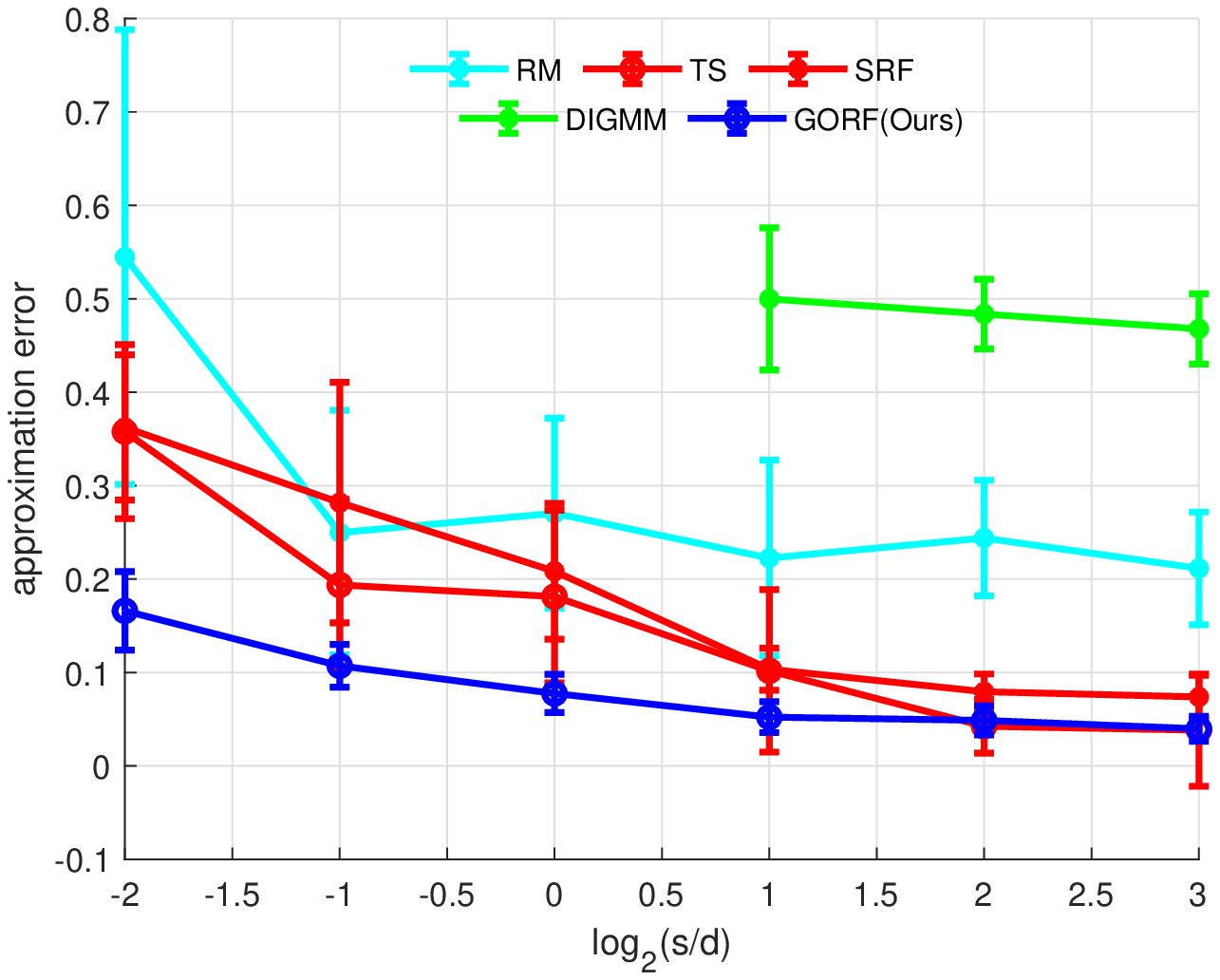}
			\end{minipage}%
		}%
		\subfigure{
			\begin{minipage}[t]{0.3\linewidth}
				\centering
				\includegraphics[width=2in]{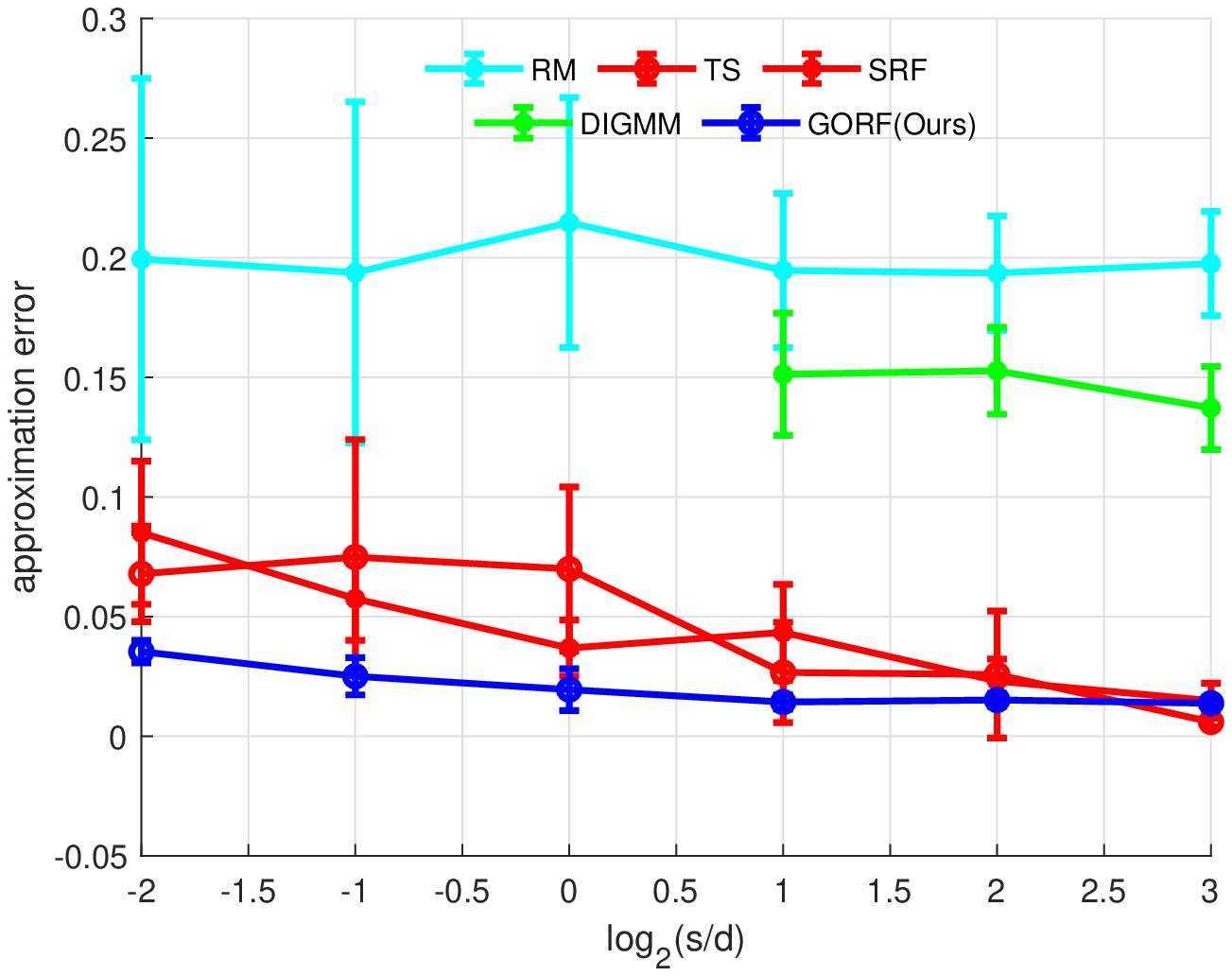}
			\end{minipage}
		}%
	}%
	
	\subfigure{
		\setcounter{subfigure}{0}
		\subfigure[$letter$]{
			\begin{minipage}[t]{0.3\linewidth}
				\centering
				\includegraphics[width=2in]{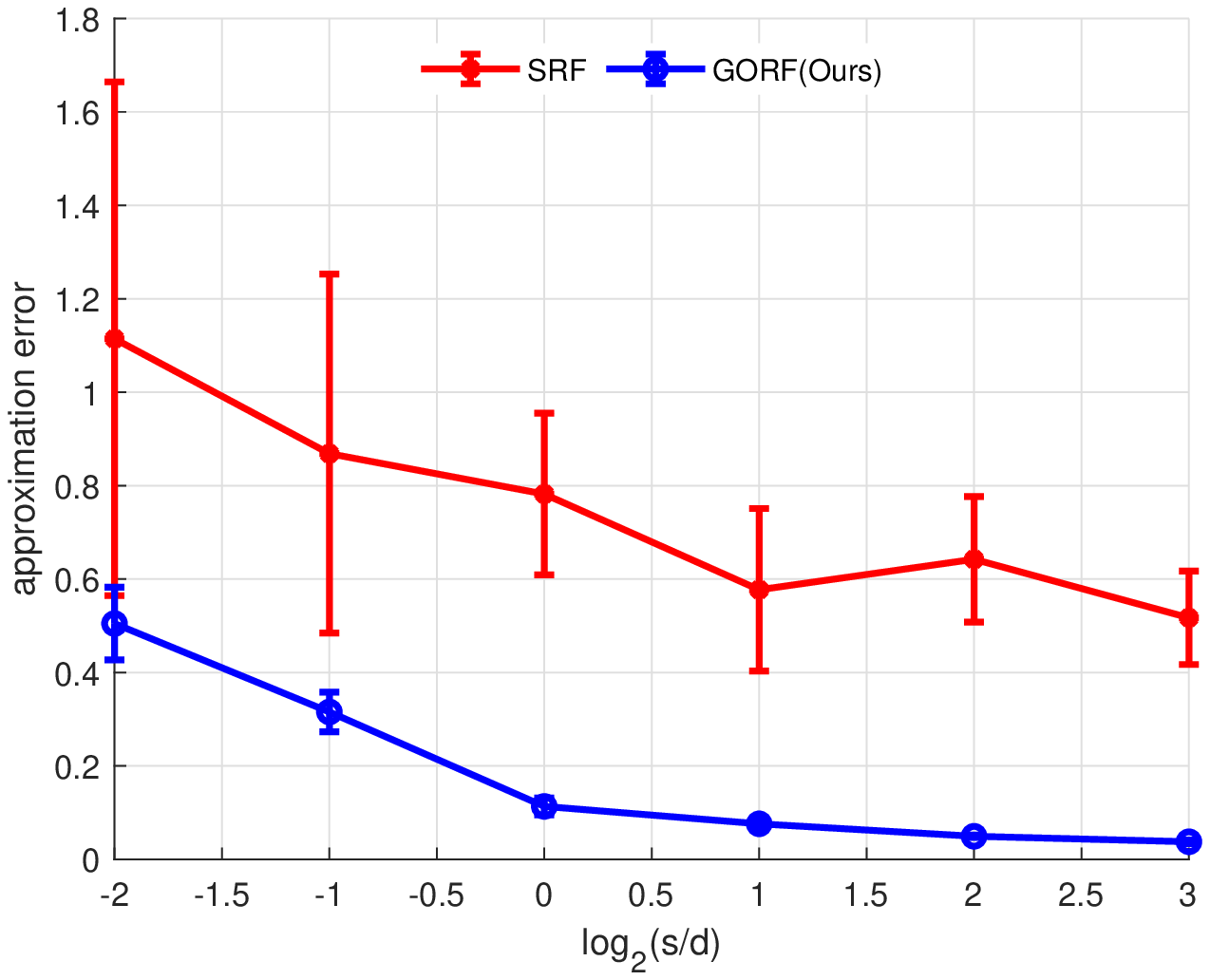}
			\end{minipage}%
		}%
		\subfigure[$ijcnn1$]{
			\begin{minipage}[t]{0.3\linewidth}
				\centering
				\includegraphics[width=2in]{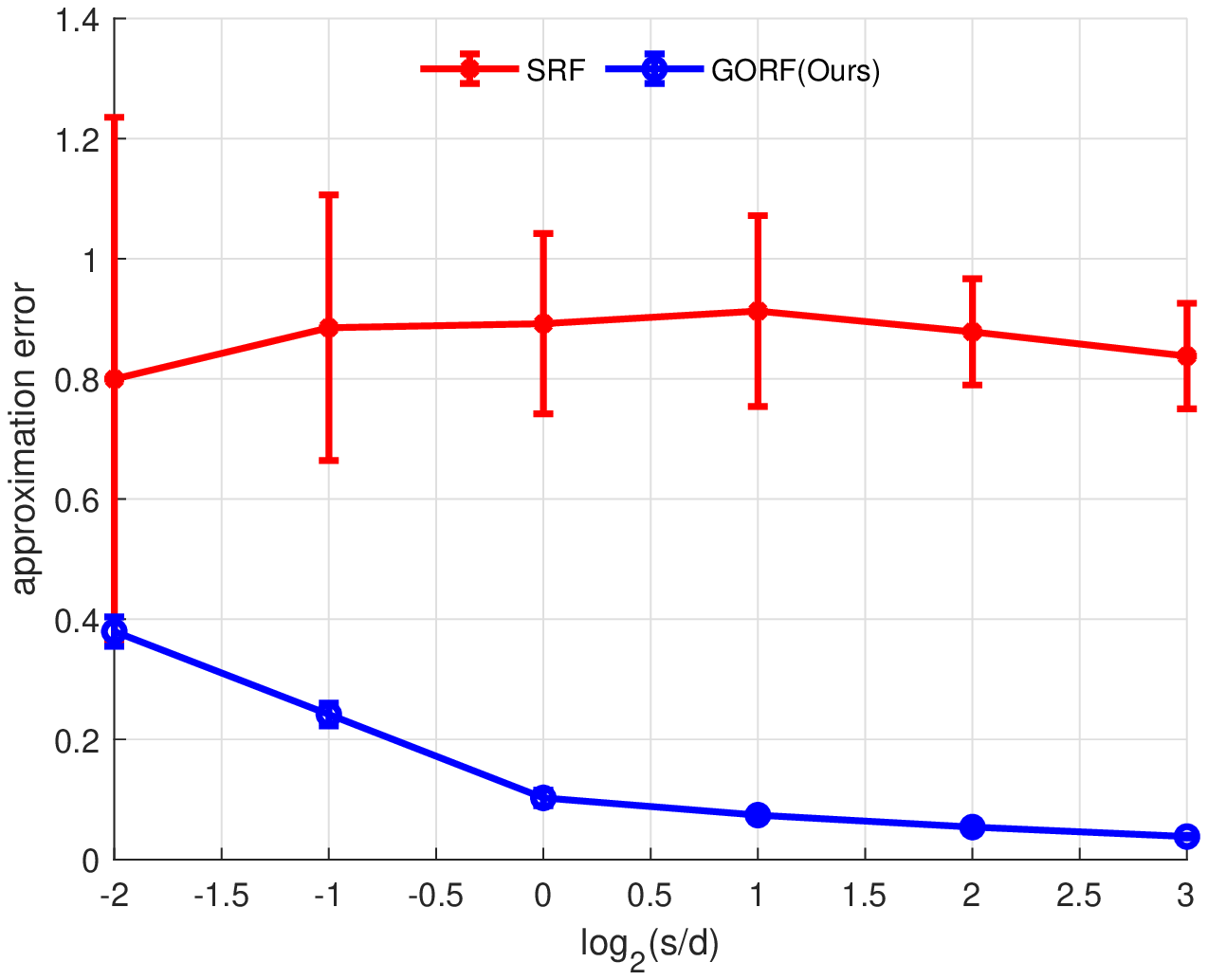}
			\end{minipage}%
		}%
		\subfigure[$usps$]{
			\begin{minipage}[t]{0.3\linewidth}
				\centering
				\includegraphics[width=2in]{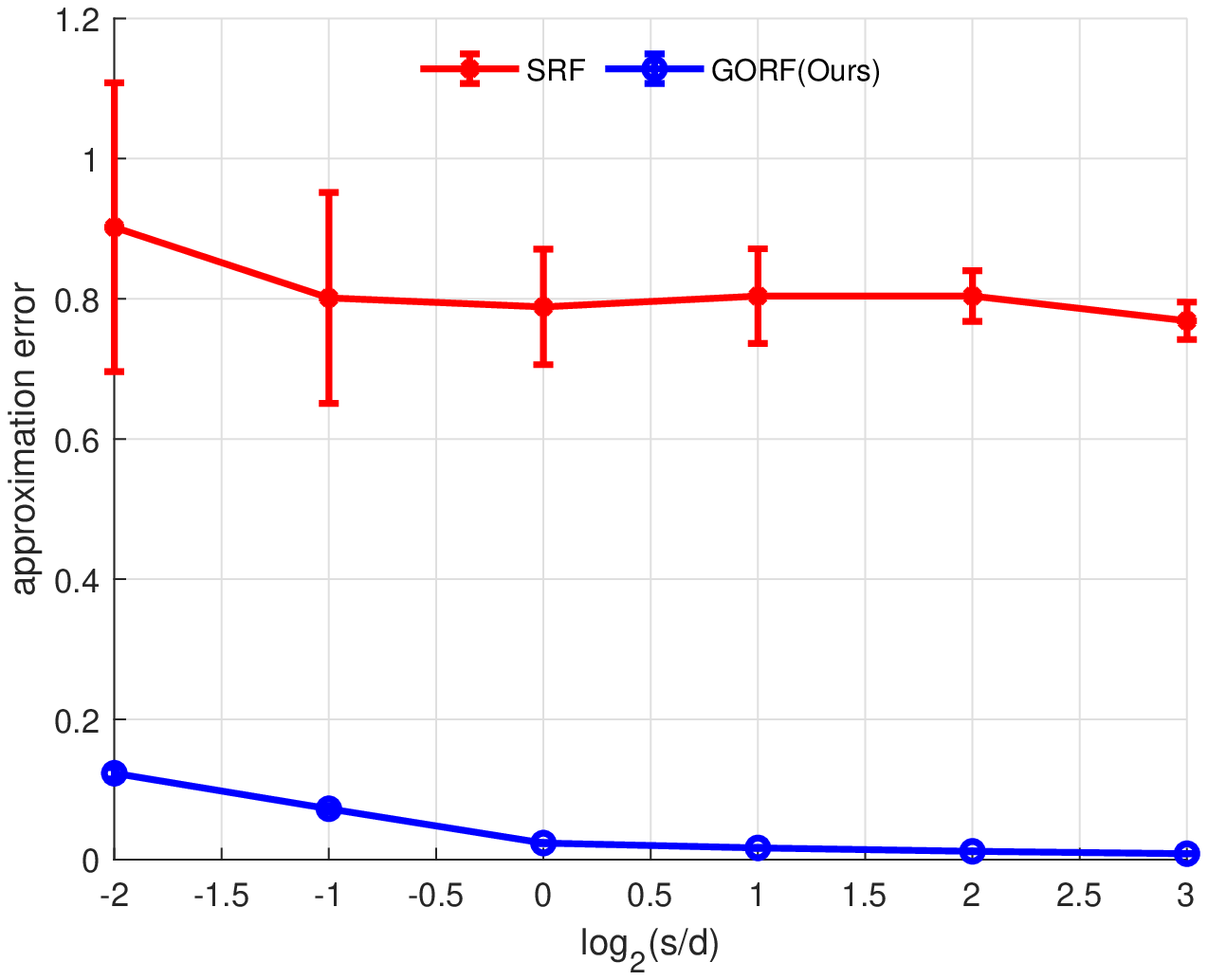}
			\end{minipage}
		}%
	}%
	\centering
	\caption{Comparisons of various algorithms for kernel approximation in terms of approximation error across two typical stationary indefinite kernels and three datasets with different dimensions. Top: polynomial kernel on the unit sphere. Below: delta-gaussian kernel}
\end{figure*}

\begin{table*}[htbp]
	\centering
	\caption{COMPARISON RESULTS BETWEEN APPLYING ORTHOGONAL SAMPLING AND I.I.D SAMPLING ON
		\label{orth}
		STATIONARY INDEFINITE KERNELS IN TERMS OF APPROXIMATION ERROR (MEAN$\pm$STD.). THE LOWEST ERROR IS HIGHLIGHTED IN \textbf{BOLDFACE}}
	\begin{center}
		\begin{tabular}{c c c c c c c}
			\hline
			\textbf{Kernel}&\textbf{DataSet}&\textbf{Method}&\textbf{s=1/2d}&\textbf{s=d}&\textbf{s=2d}&\textbf{s=8d} \\
			\hline
			\multirow{6}{*}{polynomial}&\multirow{2}{*}{$letter$}&\textit{GRFF} & 0.0859$\pm$0.0309& 0.0547$\pm$0.0078& 0.0469$\pm$0.0109& 0.0261$\pm$0.0059\\
			\multirow{6}{*}{~}&\multirow{2}{*}{~}&\textit{GORF}& \textbf{0.0716$\pm$0.0175}& \textbf{0.0495$\pm$0.0139}& \textbf{0.0360$\pm$0.0110}& \textbf{0.0231$\pm$0.0078}\\
			\cline{2-7}
			\multirow{6}{*}{~}&\multirow{2}{*}{$ijcnn1$}&\textit{GRFF} & 0.1159$\pm$0.0158& 0.0907$\pm$0.0194& 0.0794$\pm$0.0142& 0.0433$\pm$0.0059\\
			\multirow{6}{*}{~}&\multirow{2}{*}{~}&\textit{GORF}& \textbf{0.1072$\pm$0.0228}& \textbf{0.0775$\pm$0.0204}& \textbf{0.0487$\pm$0.0155}& \textbf{0.0397$\pm$0.0135}\\
			\cline{2-7}
			\multirow{6}{*}{~}&\multirow{2}{*}{$usps$}&\textit{GRFF} & 0.0270$\pm$0.0056& 0.0213$\pm$0.0063& 0.0160$\pm$0.0029& 0.0137$\pm$0.0020\\
			\multirow{6}{*}{~}&\multirow{2}{*}{~}&\textit{GORF}& \textbf{0.0251$\pm$0.0078}& \textbf{0.0194$\pm$0.0087}& \textbf{0.0143$\pm$0.0030}& \textbf{0.0137$\pm$0.0016}\\
			\cline{1-7}
			\multirow{6}{*}{delta-gaussian}&\multirow{2}{*}{$letter$}&\textit{GRFF} & 0.3918$\pm$ 0.0428& 0.2736$\pm$0.0345& 0.1887$\pm$0.0201& 0.1017$\pm$0.0088\\
			\multirow{6}{*}{~}&\multirow{2}{*}{~}&\textit{GORF}& \textbf{0.3154$\pm$0.0424}& \textbf{0.1133$\pm$0.0181}& \textbf{0.0760$\pm$0.0090}& \textbf{0.0376$\pm$0.0039}\\
			\cline{2-7}
			\multirow{6}{*}{~}&\multirow{2}{*}{$ijcnn1$}&\textit{GRFF} & 0.2924$\pm$0.0188& 0.2171$\pm$0.0222& 0.1504$\pm$0.0134& 0.0757$\pm$0.0081\\
			\multirow{6}{*}{~}&\multirow{2}{*}{~}&\textit{GORF}& \textbf{0.2415$\pm$0.0190}& \textbf{0.1026$\pm$0.0129}& \textbf{0.0739$\pm$0.0065}& \textbf{0.0383$\pm$0.0022}\\
			\cline{2-7}
			\multirow{6}{*}{~}&\multirow{2}{*}{$usps$}&\textit{GRFF} & 0.1005$\pm$0.0061& 0.0690$\pm$0.0050& 0.0500$\pm$0.0024& 0.0253$\pm$0.0023\\
			\multirow{6}{*}{~}&\multirow{2}{*}{~}&\textit{GORF}& \textbf{0.0724$\pm$0.0049}& \textbf{0.0235$\pm$0.0009}& \textbf{0.0166$\pm$0.0008}& \textbf{0.0083$\pm$0.0003}\\
			\hline
		\end{tabular}
		\label{tab1}
	\end{center}
\end{table*}

\subsection{Approximation Error}
Approximation error is the most important issue for RFF methods. Unbiasedness and variance determine the overall approximation error. We compare our method with other existing methods to demonstrate the superiority of reducing approximation error in this part and investigate the effect of orthogonality on reduction of approximation error for ablation study.

In this part, to quantify the approximation error, relative error $\frac{||K-\hat{K}||_{F}}{||K||_{F}}$ is calculated  where $K$ and $\hat{K}$ denote the exact kernel matrix and and its approximated kernel matrix on 1000 randomly selected samples. The parameters in polynomial kernel on the unit sphere and delta-gaussian kernel are the same as that in Sec. \ref{sec::var_redcut}. The approximation error is represented by mean and standard deviation in 10 repetitive experiments. 
\subsubsection{comparison with existing methods}
In this part, several kernel approximation baseline methods are utilized for comparison in terms of approximation error, including Random Maclaurin (RM) \cite{pmlr-v22-kar12}, Tensor Sketch (TS) \cite{TensorSketch}, Spherical Random Features (SRF) \cite{NIPS2015_5943} and Double-Infinite Gaussian Mixtures Model (DIGMM) \cite{Liu8830377}. RM and TS are the unbiased estimations for polynomial kernel. SRF and DIGMM are two biased approximation for arbitrary stationary kernels using the Gaussian Mixture Model (GMM).

The comparison between our GORF method and other existing method in terms of approximation ability is demonstrated in Fig. 2. It shows that our method achieves lowest approximation error. The lowest approximation error is attributed to two important factors: 1) unbiasness. Our GORF method is unbiased compared with the current stationary indefinite kernel

\begin{figure*}[htbp]
	\centering
	\subfigure{
		\subfigure{
			\begin{minipage}[t]{0.3\linewidth}
				\centering
				\includegraphics[width=2in]{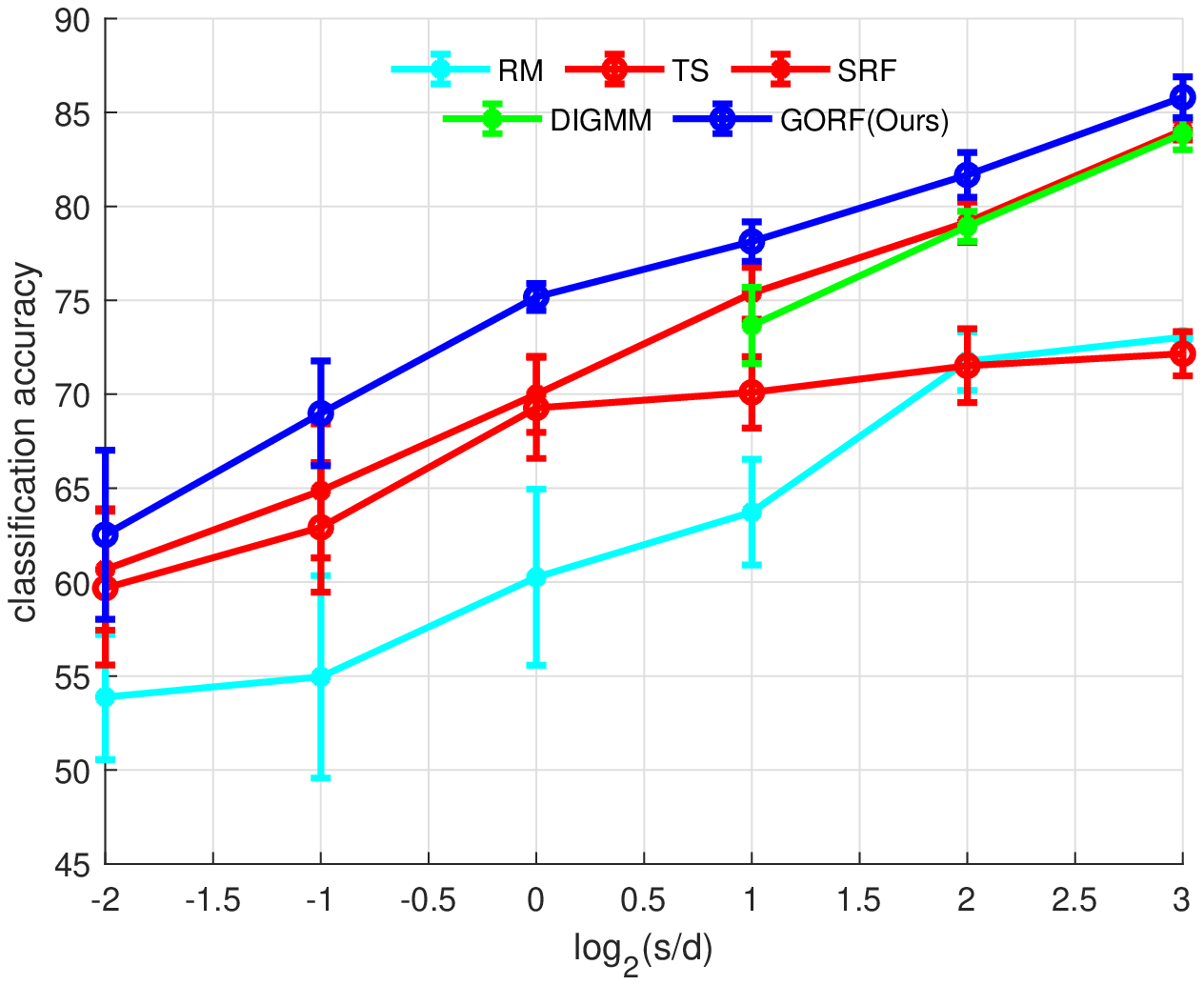}
			\end{minipage}%
		}%
		\subfigure{
			\begin{minipage}[t]{0.3\linewidth}
				\centering
				\includegraphics[width=2in]{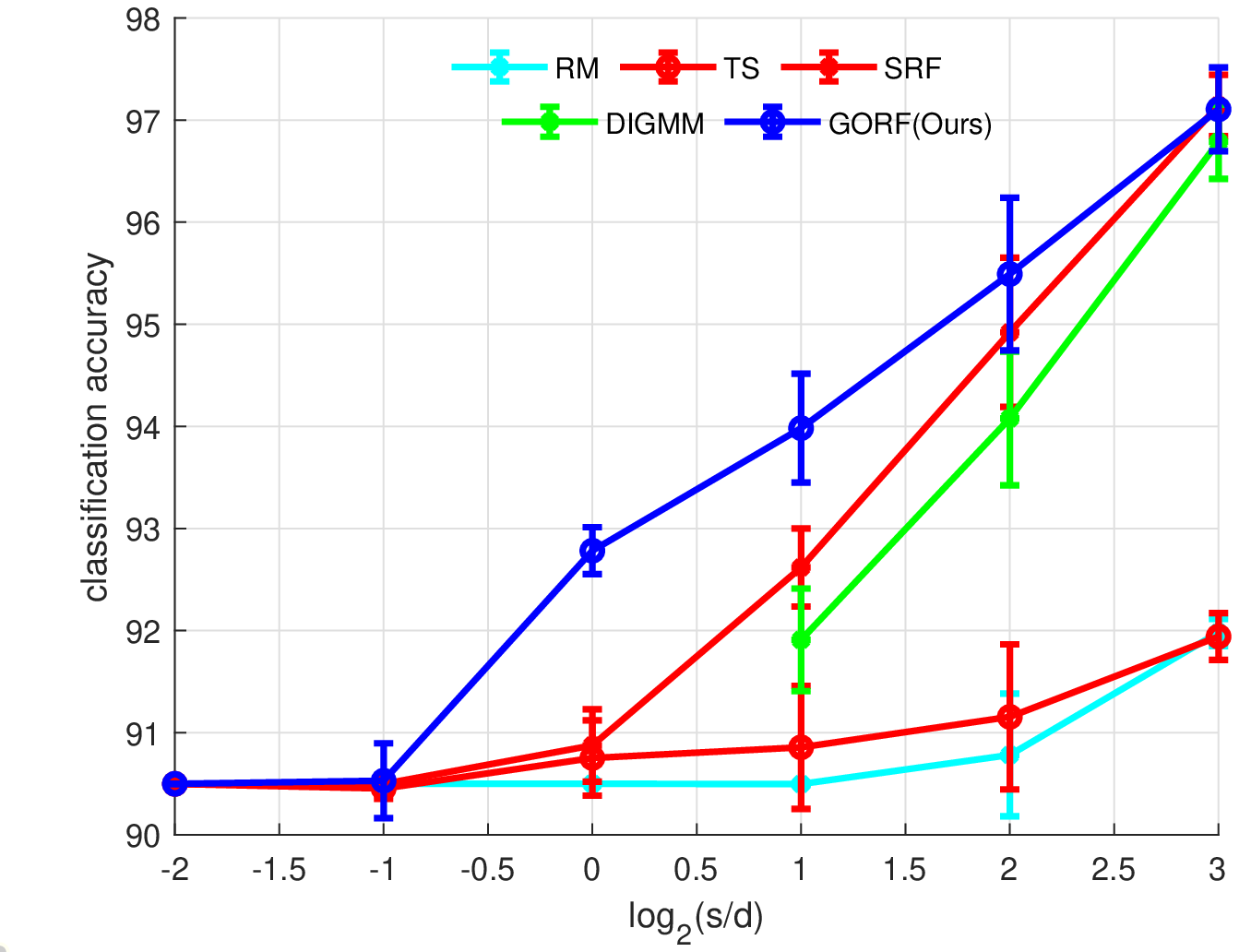}
			\end{minipage}%
		}%
		\subfigure{
			\begin{minipage}[t]{0.3\linewidth}
				\centering
				\includegraphics[width=2in]{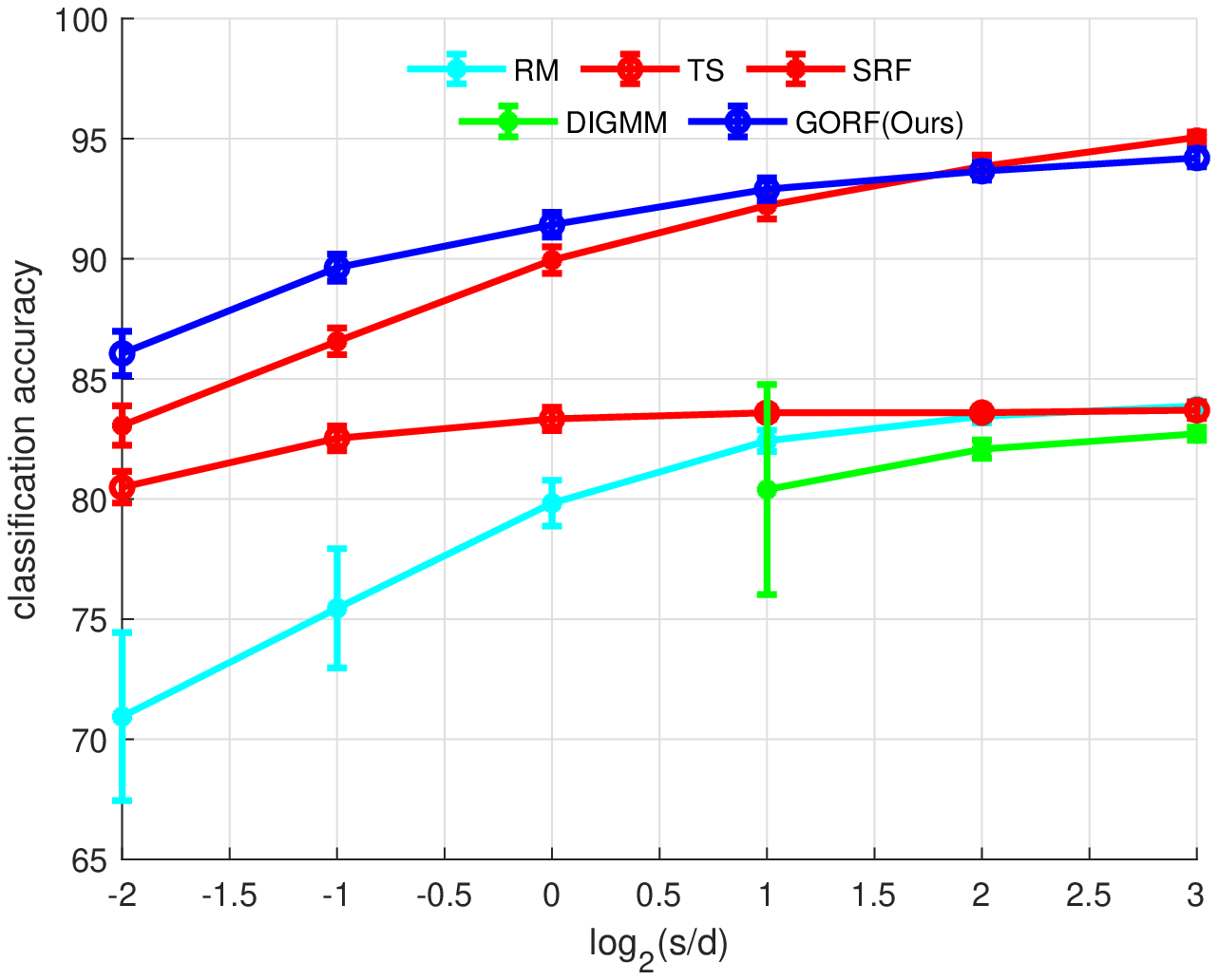}
			\end{minipage}
		}%
	}%
	
	\subfigure{
		\setcounter{subfigure}{0}
		\subfigure[$letter$]{
			\begin{minipage}[t]{0.3\linewidth}
				\centering
				\includegraphics[width=2in]{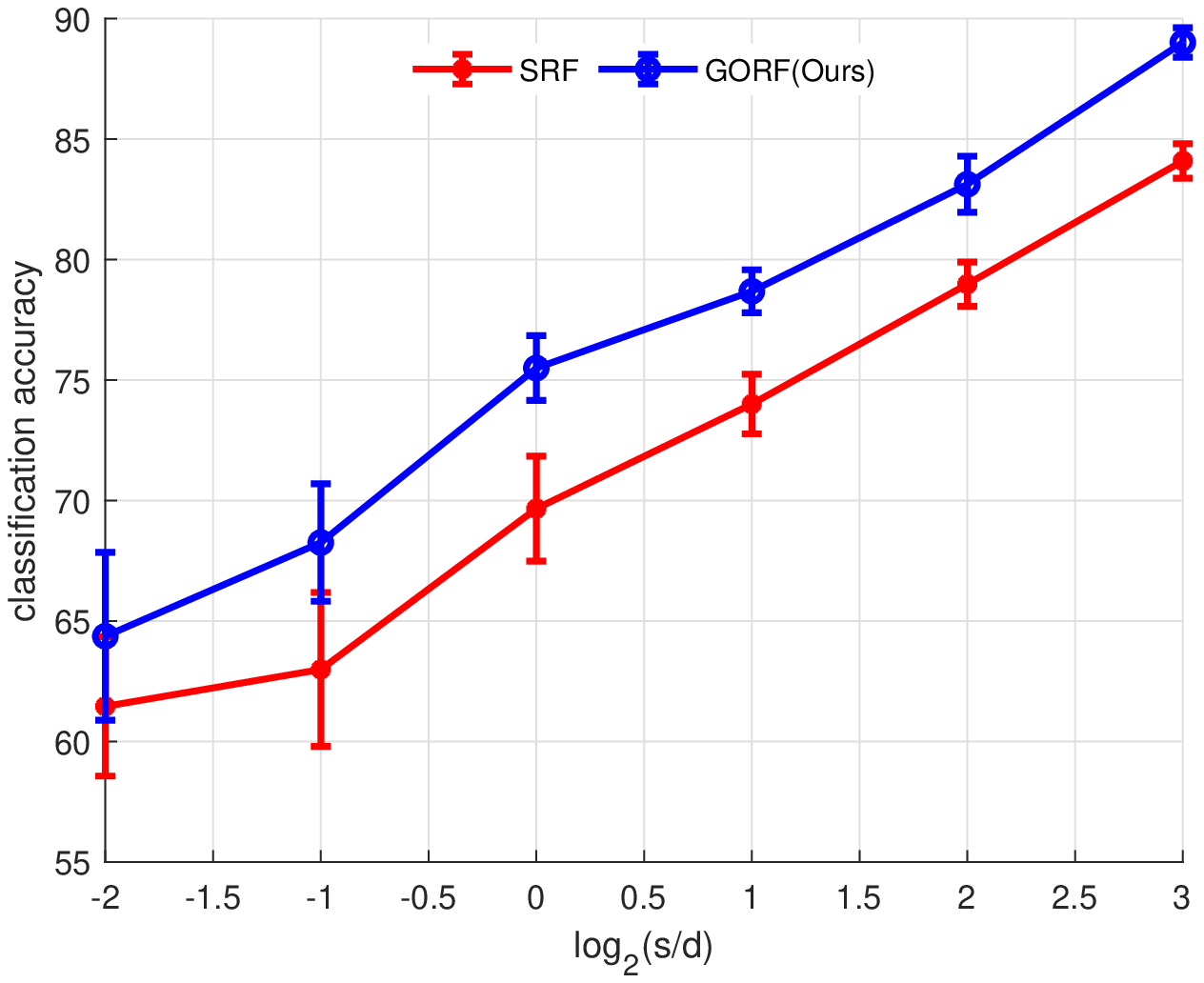}
			\end{minipage}%
		}%
		\subfigure[$ijcnn1$]{
			\begin{minipage}[t]{0.3\linewidth}
				\centering
				\includegraphics[width=2in]{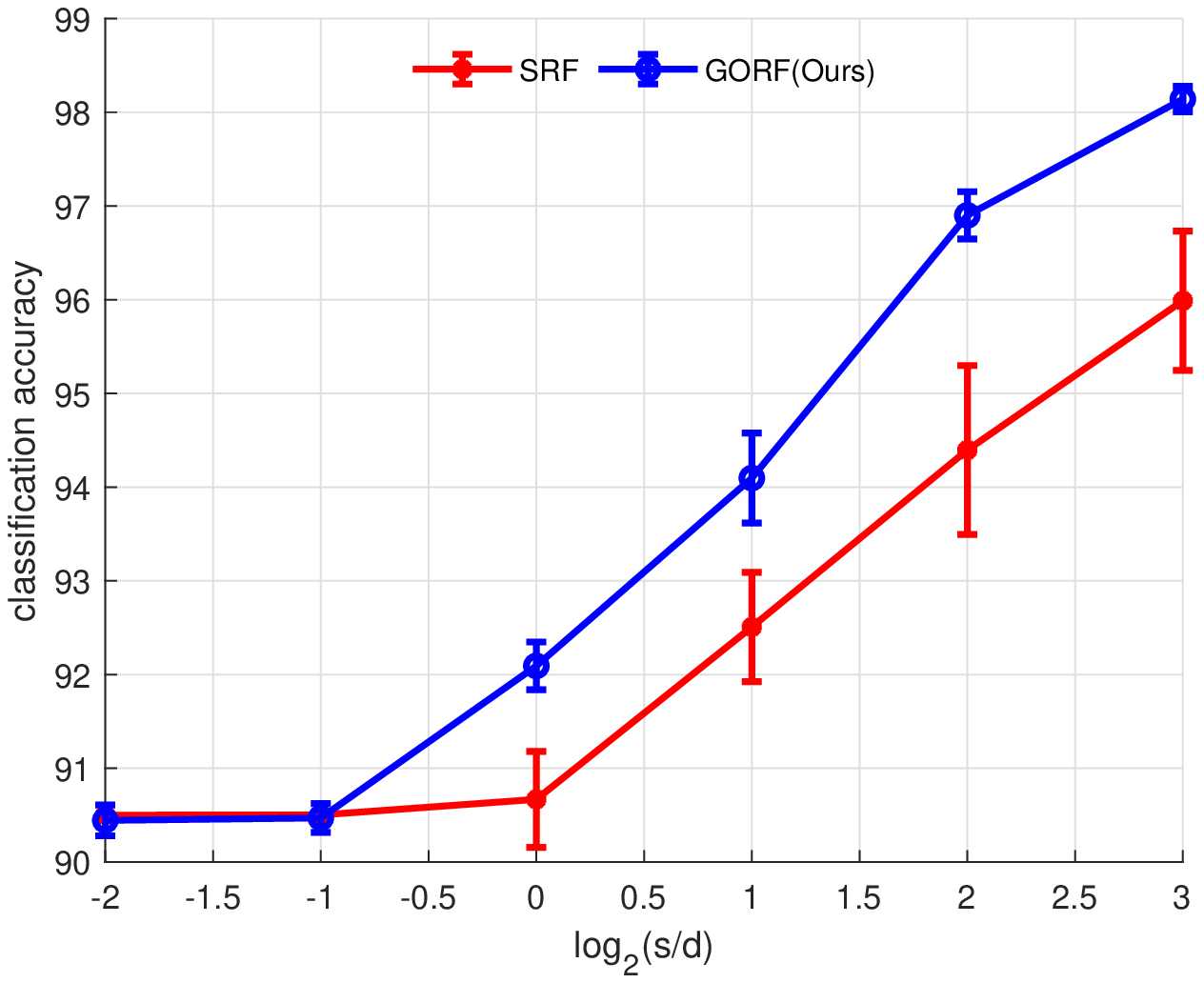}
			\end{minipage}%
		}%
		\subfigure[$usps$]{
			\begin{minipage}[t]{0.3\linewidth}
				\centering
				\includegraphics[width=2in]{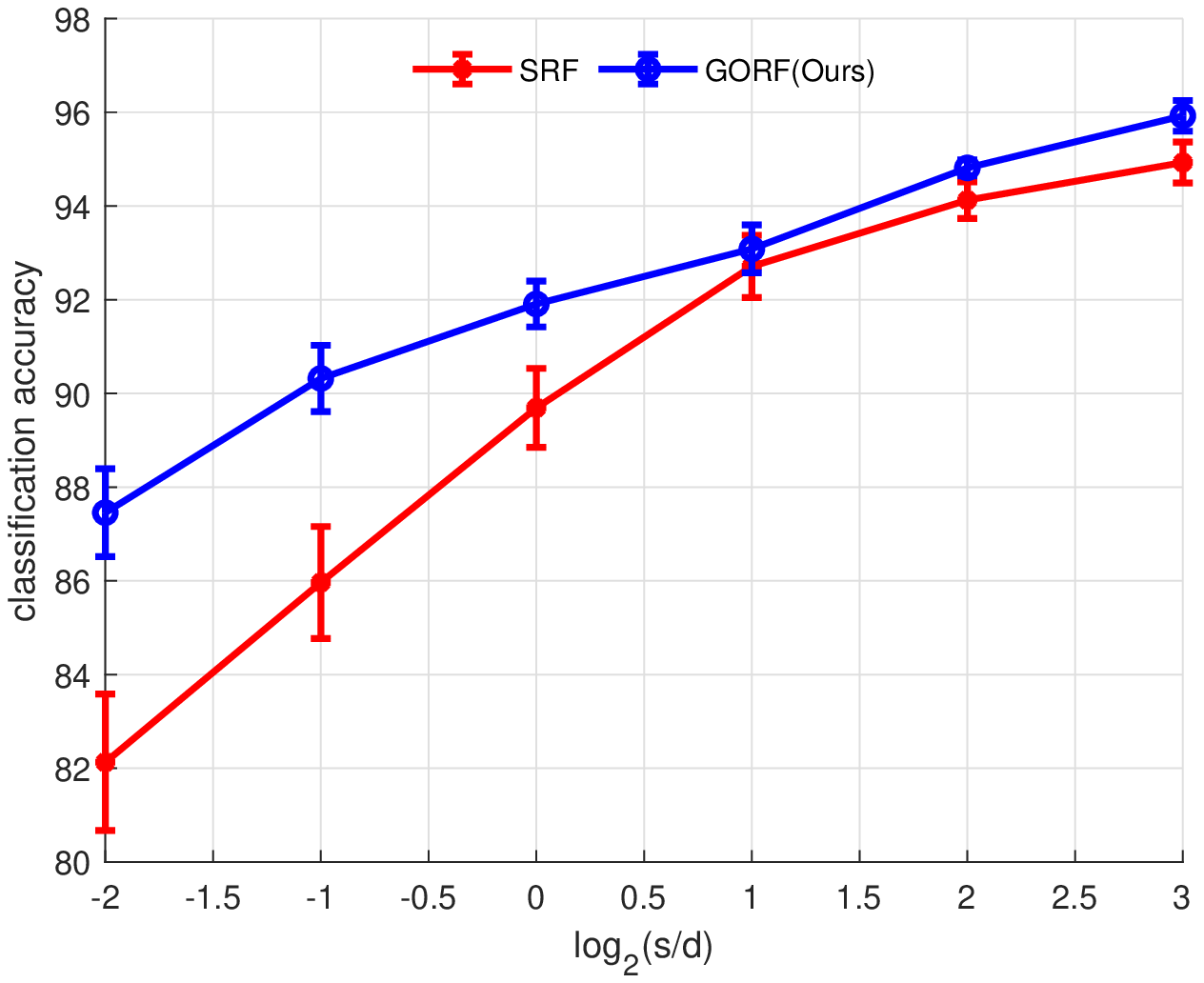}
			\end{minipage}
		}%
	}%
	\centering
	\caption{Comparisons of various algorithms for SVM classification task in terms of accuracy across two typical stationary indefinite kernels and three datasets with different dimensions. Top: polynomial kernel on the unit sphere. Below: delta-gaussian kernel}
\end{figure*}

\noindent approximation methods, like SRF and DIGMM. Especially for delta-gaussian kernel whose positive and negative measure are almost the same, our GORF method retains both of these two parts but SRF method simply retains either the positive or the negative part, resulting a significant drop in approximation error for our method. 2) low variance. Our method achieves lower variance based on the unbiasedness. Our GORF method achieves faster convergence rate than other unbiased methods like RM and TS in polynomial kernel approximation because of low variance brought by orthogonality and RFF. Notice that for the baseline method DIGMM, it requires $s\geq2d$ because of the intolerance of the non-PD matrices in Cholosky decomposition and the complex value in polygamma function computation occurred in $s\textless 2d$.

\subsubsection{effect of orthogonality}
We evaluate orthogonality, the crucial part in GORF algorithm, in terms of approximation error for further ablation study in this part . Table \ref{orth} demonstrates that orthogonal
sampling brings a further substantial approximation error reduction which is averagely above 10$\%$ drop because of the significant decrease in variance. For delta-gaussian kernel, the decrease in approximation error becomes more significant, around 20$\%$ drop.

\subsection{Performance in classification and regression}
We compare our GORF method and other existing methods when considering the specific support vector classification and regression task. In classification experiment, datasets $letter$, $ijcnn1$ and $usps$ are utilized. and classification accuracy is the evaluation metric. In regression experiment, the dataset $housing$ is used and root mean square error (RMSE) between ground truth and predictions is utilized as evaluation metric. We choose $liblinear$, a toolkit for support vector machine and support vector regression in our experiment. The parameter $C$ in $liblinear$ \cite{10.5555/1390681.1442794} is set as 1000. All the other hyperparameters is the same as that in the experiments above.

In the classification experiment, we could observe that our GORF method achieves better classification accuracy than the existing stationary indefinite kernel approximation methods for almost all the number of random features in Fig. 3. In the regression task, Table \ref{result::poly} and \ref{result::delta} demonstrates that our GORF method achieves lower RMSE and possesses better ability to predict the housing prices in testing set. These results could be attributed to better approximation to the originally selected kernels. Better approximation contributes to better performance in classification and regression. And unbiasedness and lower variance are the main factors of lower approximation error in our GORF algorithm.
\begin{table}[htbp]
\caption{REGRESSION ERROR FOR KERNEL APPROXIMATION METHODS ON POLYNOMIAL KERNEL AND $HOUSING$ DATASET (RMSE: MEAN$\pm$STD). THE LOWEST ERROR IS HIGHLIGHTED IN \textbf{BOLDFACE}}
\label{regression}
\begin{center}
\begin{tabular}{c c c c c}
\hline
\textbf{Methods}&\textbf{s=2d}&\textbf{s=4d}&\textbf{s=8d}\\
\cline{1-4}
RM & 7.153$\pm$1.772 & 5.436$\pm$0.917 & 4.491$\pm$0.008\\
TS & 5.414$\pm$0.879 & 4.772$\pm$0.177 & 4.657$\pm$0.316\\
SRF & 4.391$\pm$0.368 & 3.906$\pm$0.219 & 3.555$\pm$0.130\\
DIGMM & 4.897$\pm$0.368 & 4.130$\pm$0.324 & 4.000$\pm$0.475\\
GORF(OURS) & \textbf{4.079$\pm$0.233} & \textbf{3.817$\pm$0.204} & \textbf{3.472$\pm$0.137}\\
\hline
\end{tabular}
\label{result::poly}
\end{center}
\end{table}

\section{Conclusion}
In this paper, we propose generalized orthogonal random features, an unbiased estimation with lower variance for stationary indefinite kernel approximation. We theoretically verify the unbiasedness and derive the variance reduction. Experimental results show that our proposed algorithm achieves lower variance and approximation error. In specific support vector classification and regression problem, our method obtains better classification and regression ability compared with the existing methods.

\begin{table}[htbp]
\caption{REGRESSION ERROR FOR KERNEL APPROXIMATION METHODS ON DELTA-GAUSSIAN KERNEL AND $HOUSING$ DATASET (RMSE: MEAN$\pm$STD). THE LOWEST ERROR IS HIGHLIGHTED IN \textbf{BOLDFACE}}
\label{regression}
\begin{center}
\begin{tabular}{c c c c c}
\hline
\textbf{Methods}&\textbf{s=2d}&\textbf{s=4d}&\textbf{s=8d}\\
\cline{1-4}
SRF & 5.432$\pm$0.729 & 3.845$\pm$0.379 & 3.321$\pm$0.274\\
GORF(OURS) & \textbf{3.739$\pm$0.360} & \textbf{3.474$\pm$0.330} & \textbf{3.164$\pm$0.452}\\
\hline
\end{tabular}
\label{result::delta}
\end{center}
\end{table}

\section*{Acknowledgment}
This work is supported by National Key Research Development Project (No. 2018AAA0100702) and National Natural Science Foundation of China (Nos. 61977046, 61876107, U1803261) and Committee of Science of Technology, Shanghai, China (No. 19510711200).

\bibliographystyle{IEEEtran}
\bibliography{IEEEabrv, refs}

\begin{thebibliography}{10}
\providecommand{\url}[1]{#1}
\csname url@samestyle\endcsname
\providecommand{\newblock}{\relax}
\providecommand{\bibinfo}[2]{#2}
\providecommand{\BIBentrySTDinterwordspacing}{\spaceskip=0pt\relax}
\providecommand{\BIBentryALTinterwordstretchfactor}{4}
\providecommand{\BIBentryALTinterwordspacing}{\spaceskip=\fontdimen2\font plus
\BIBentryALTinterwordstretchfactor\fontdimen3\font minus
  \fontdimen4\font\relax}
\providecommand{\BIBforeignlanguage}[2]{{%
\expandafter\ifx\csname l@#1\endcsname\relax
\typeout{** WARNING: IEEEtran.bst: No hyphenation pattern has been}%
\typeout{** loaded for the language `#1'. Using the pattern for}%
\typeout{** the default language instead.}%
\else
\language=\csname l@#1\endcsname
\fi
#2}}
\providecommand{\BIBdecl}{\relax}
\BIBdecl

\bibitem{1995Support}
C.~Cortes and V.~N. Vapnik, ``Support-vector networks,'' \emph{Machine
  Learning}, vol.~20, no.~3, pp. 273--297, 1995.

\bibitem{Learningkernels}
B.~Schlkopf, A.~J. Smola, and F.~Bach, \emph{Learning with Kernels: Support
  Vector Machines, Regularization, Optimization, and Beyond}.\hskip 1em plus
  0.5em minus 0.4em\relax The MIT Press, 2018.

\bibitem{1997Support}
H.~Drucker, C.~J.~C. Burges, L.~Kaufman, J.~C. Chris, B.~L. Kaufman, A.~Smola,
  and V.~Vapnik, ``Support vector regression machines,'' \emph{Advances in
  Neural Information Processing Systems}, vol.~28, no.~7, pp. 779--784, 1997.

\bibitem{pmlr-v28-wilson13}
A.~Wilson and R.~Adams, ``Gaussian process kernels for pattern discovery and
  extrapolation,'' in \emph{Proceedings of the 30th International Conference on
  Machine Learning}, Atlanta, Georgia, USA, 17--19 Jun 2013, pp. 1067--1075.

\bibitem{Lazaro2010Sparse}
M.~Lazaro-Gredilla, J.~Quinonero-Candela, C.~E. Rasmussen, and A.~R.
  Figueiras-Vidal, ``Sparse spectrum gaussian process regression,''
  \emph{Journal of Machine Learning Research}, vol.~11, pp. 1865--1881, 2010.

\bibitem{1997Kernel}
B.~Scholkopf and A.~Smola, ``Kernel principal component analysis,'' in
  \emph{International Conference on Artificial Neural Networks}, 1997.

\bibitem{Carlos2018Convex}
C.~M. Alaíz, M.~Fanuel, and J.~A.~K. Suykens, ``Convex formulation for kernel
  pca and its use in semisupervised learning,'' \emph{IEEE Transactions on
  Neural Networks and Learning Systems}, vol.~29, no.~8, pp. 3863--3869, 2018.

\bibitem{NIPS2007_3182}
A.~Rahimi and B.~Recht, ``Random features for large-scale kernel machines,'' in
  \emph{Advances in Neural Information Processing Systems 20}, 2008, pp.
  1177--1184.

\bibitem{pmlr-v28-le13}
Q.~Le, T.~Sarlos, and A.~Smola, ``Fastfood - computing hilbert space expansions
  in loglinear time,'' ser. Proceedings of Machine Learning Research, vol.~28,
  no.~3, Atlanta, Georgia, USA, 17--19 Jun 2013, pp. 244--252.

\bibitem{pmlr-v32-yangb14}
J.~Yang, V.~Sindhwani, H.~Avron, and M.~Mahoney, ``Quasi-monte carlo feature
  maps for shift-invariant kernels,'' ser. Proceedings of Machine Learning
  Research, vol.~32, no.~1, Bejing, China, 22--24 Jun 2014, pp. 485--493.

\bibitem{Circulant}
C.~Feng, Q.~Hu, and S.~Liao, ``Random feature mapping with signed circulant
  matrix projection,'' in \emph{Proceedings of the 24th International
  Conference on Artificial Intelligence}, ser. IJCAI'15, 2015, p. 3490–3496.

\bibitem{NIPS2016_6246}
F.~X. Yu, A.~T. Suresh, K.~M. Choromanski, D.~N. Holtmann-Rice, and S.~Kumar,
  ``Orthogonal random features,'' in \emph{Advances in Neural Information
  Processing Systems 29}, 2016, pp. 1975--1983.

\bibitem{pmlr-v84-choromanski18a}
K.~Choromanski, M.~Rowland, T.~Sarlos, V.~Sindhwani, R.~Turner, and A.~Weller,
  ``The geometry of random features,'' ser. Proceedings of Machine Learning
  Research, vol.~84, Playa Blanca, Lanzarote, Canary Islands, 09--11 Apr 2018,
  pp. 1--9.

\bibitem{pmlr-v97-choromanski19a}
K.~Choromanski, M.~Rowland, W.~Chen, and A.~Weller, ``Unifying orthogonal
  {M}onte {C}arlo methods,'' ser. Proceedings of Machine Learning Research,
  vol.~97, Long Beach, California, USA, 09--15 Jun 2019, pp. 1203--1212.

\bibitem{pmlr-v54-bojarski17a}
M.~Bojarski, A.~Choromanska, K.~Choromanski, F.~Fagan, C.~Gouy-Pailler,
  A.~Morvan, N.~Sakr, T.~Sarlos, and J.~Atif, ``{Structured adaptive and random
  spinners for fast machine learning computations},'' ser. Proceedings of
  Machine Learning Research, vol.~54, Fort Lauderdale, FL, USA, 20--22 Apr
  2017, pp. 1020--1029.

\bibitem{NIPS2000_1790}
A.~J. Smola, Z.~L. \'{O}v\'{a}ri, and R.~C. Williamson, ``Regularization with
  dot-product kernels,'' in \emph{Advances in Neural Information Processing
  Systems 13}, 2001, pp. 308--314.

\bibitem{NIPS2018_8076}
A.~Jacot, F.~Gabriel, and C.~Hongler, ``Neural tangent kernel: Convergence and
  generalization in neural networks,'' in \emph{Advances in Neural Information
  Processing Systems 31}, 2018, pp. 8571--8580.

\bibitem{pmlr-v80-oglic18a}
D.~Oglic and T.~Gaertner, ``Learning in reproducing kernel krein spaces,'' ser.
  Proceedings of Machine Learning Research, vol.~80, Stockholmsmässan,
  Stockholm Sweden, 10--15 Jul 2018, pp. 3859--3867.

\bibitem{Xiaolin2017Classification}
Xiaolin, Huang, A.~K. Johan, Suykens, Shuning, Wang, Joachim, Hornegger,
  Andreas, and Maier, ``Classification with truncated $\ell_{1}$ distance
  kernel.'' \emph{IEEE transactions on neural networks and learning systems},
  2017.

\bibitem{NIPS2015_5943}
J.~Pennington, F.~X.~X. Yu, and S.~Kumar, ``Spherical random features for
  polynomial kernels,'' in \emph{Advances in Neural Information Processing
  Systems 28}, 2015, pp. 1846--1854.

\bibitem{pmlr-v22-kar12}
P.~Kar and H.~Karnick, ``Random feature maps for dot product kernels,'' ser.
  Proceedings of Machine Learning Research, vol.~22, La Palma, Canary Islands,
  21--23 Apr 2012, pp. 583--591.

\bibitem{TensorSketch}
\BIBentryALTinterwordspacing
N.~Pham and R.~Pagh, ``Fast and scalable polynomial kernels via explicit
  feature maps,'' in \emph{Proceedings of the 19th ACM SIGKDD International
  Conference on Knowledge Discovery and Data Mining}, ser. KDD ’13, 2013, p.
  239–247. [Online]. Available: \url{https://doi.org/10.1145/2487575.2487591}
\BIBentrySTDinterwordspacing

\bibitem{2012Efficient}
A.~Vedaldi and A.~Zisserman, ``Efficient additive kernels via explicit feature
  maps,'' vol.~34, no.~3, 2012, pp. 480--492.

\bibitem{2017CROification}
M.~Kafai and K.~Eshghi, ``Croification: Accurate kernel classification with the
  efficiency of sparse linear svm,'' 2017, pp. 1--1.

\bibitem{Liu8830377}
F.~{Liu}, X.~{Huang}, L.~{Shi}, J.~{Yang}, and J.~A.~K. {Suykens}, ``A
  double-variational bayesian framework in random fourier features for
  indefinite kernels,'' \emph{IEEE Transactions on Neural Networks and Learning
  Systems}, vol.~31, no.~8, pp. 2965--2979, 2020.

\bibitem{Liu2020GeneralizingRF}
F.~Liu, X.~Huang, Y.~Chen, and J.~Suykens, ``Fast learning in reproducing
  kernel krein spaces via signed measures,'' in \emph{Proceedings of The 24th
  International Conference on Artificial Intelligence and Statistics}, vol.
  130, 13--15 Apr 2021, pp. 388--396.

\bibitem{bochner}
S.~Bochner, ``Harmonic analysis and theory of probability,'' \emph{Physics
  Today}, vol.~9, 01 1956.

\bibitem{NEURIPS2018_6e923226}
M.~Munkhoeva, Y.~Kapushev, E.~Burnaev, and I.~Oseledets, ``Quadrature-based
  features for kernel approximation,'' in \emph{Advances in Neural Information
  Processing Systems}, vol.~31, 2018, pp. 9147--9156.

\bibitem{Athreya2006MeasureTA}
K.~Athreya and S.~Lahiri, ``Measure theory and probability theory.''\hskip 1em
  plus 0.5em minus 0.4em\relax Springer, 2006.

\bibitem{Kubrusly2015EssentialsOM}
C.~S. Kubrusly, ``Essentials of measure theory.''\hskip 1em plus 0.5em minus
  0.4em\relax Springer, 2015.

\bibitem{NonPositiveKernels}
C.~S. Ong, X.~Mary, S.~Canu, and A.~J. Smola, ``Learning with non-positive
  kernels,'' in \emph{Proceedings of the Twenty-First International Conference
  on Machine Learning}, ser. ICML ’04, 2004, p.~81.

\bibitem{10.5555/1390681.1442794}
R.-E. Fan, K.-W. Chang, C.-J. Hsieh, X.-R. Wang, and C.-J. Lin, ``Liblinear: A
  library for large linear classification,'' \emph{J. Mach. Learn. Res.},
  vol.~9, p. 1871–1874, Jun. 2008.

\end{thebibliography}

\end{document}